\documentclass[11pt]{article}

\usepackage[final]{acl}

\usepackage{multirow}
\usepackage{algorithmic}

\usepackage{adjustbox}
\usepackage{natbib}
\usepackage{algorithm}
\usepackage{algorithmic}
\usepackage{makecell}
\usepackage{booktabs}
\usepackage{colortbl}
\definecolor{lightgreen}{RGB}{220,255,220} 
\definecolor{lightred}{RGB}{255,200,200}
\usepackage{amsmath}

\usepackage{times}
\usepackage{latexsym}

\usepackage[most]{tcolorbox}

\usepackage[T1]{fontenc}
\usepackage[most]{tcolorbox}
\usepackage{xcolor}

\usepackage[utf8]{inputenc}

\usepackage{microtype}

\usepackage{inconsolata}

\usepackage{graphicx}
\newcommand{\saransh}[1]{\textcolor{black}{#1}}

\title{An Answer is just the Start: \\Related Insight Generation for Open-Ended Document-Grounded QA}

\author{
    Saransh Sharma\textsuperscript{1}, 
    Pritika Ramu\textsuperscript{2}$^{\ddagger}$, 
    Aparna Garimella\textsuperscript{1}, 
    Koyel Mukherjee\textsuperscript{1} \\[0.5em]
    \textsuperscript{1}Adobe Research, India \\
    \textsuperscript{2}University of Maryland, College Park \\[0.5em]
    \texttt{\{sarsharma, garimell, komukher\}@adobe.com}, \texttt{pramu@umd.edu}
}

\begin{document}
\maketitle
\def\thefootnote{‡}\footnotetext{Work done while at Adobe Research}\def\thefootnote{\arabic{footnote}}
\begin{abstract}
Answering open-ended questions remains challenging for AI systems because it requires synthesis, judgment, and exploration beyond factual retrieval, and users often refine answers through multiple iterations rather than accepting a single response. Existing QA benchmarks do not explicitly support this refinement process. To address this gap, we introduce a new task, \emph{\textbf{document-grounded related insight generation}}, where the goal is to generate additional insights from a document collection that help improve, extend, or rethink an initial answer to an open-ended question, ultimately supporting richer user interaction and a better overall question answering experience. 
We curate and release \textbf{SCOpE-QA} (\textbf{S}cientific \textbf{C}ollections for \textbf{Op}en-\textbf{E}nded \textbf{QA}), a dataset of 3,000 open-ended questions across 20 research collections. 
We present \textsc{InsightGen}, a two-stage approach that first constructs a thematic representation of the document collection using clustering, and then selects related context based on neighborhood selection from the thematic graph to generate diverse and relevant insights using LLMs. Extensive evaluation on 3,000 questions using two generation models and two evaluation settings shows that \textsc{InsightGen} consistently produces useful, relevant, and actionable insights, establishing a strong baseline for this new task.

\end{abstract}

\section{Introduction}
Recent advances in generative language modeling \cite{openai2023gpt4,deepmind2023gemini,bai2022constitutional,touvron2023llama} have led to a variety of AI-powered document interaction tools, including NotebookLM,\footnote{\url{https://notebooklm.google}}
 Acrobat's AI Assistant,\footnote{\url{https://www.adobe.com/acrobat/generative-ai-pdf.html}}
 and ChatPDF.\footnote{\url{https://www.chatpdf.com}}
 These applications allow users to interact with documents through question answering or to generate new content grounded in the documents. Document-grounded question answering (QA) is largely powered by retrieval-augmented generation (RAG) techniques \cite{sharma2025retrievalaugmentedgenerationcomprehensivesurvey}, which enhance LLM performance in diverse scenarios, including single-hop and multi-hop questions \cite{wu2025composeragmodularcomposablerag,li2025treehopgeneratefilterquery,zhang2025levelragenhancingretrievalaugmentedgeneration}, and multimodal questions \cite{mei2025surveymultimodalretrievalaugmentedgeneration,abootorabi2025askmodalitycomprehensivesurvey}.

\begin{figure}[t]
\centering
  \includegraphics[width=\columnwidth]{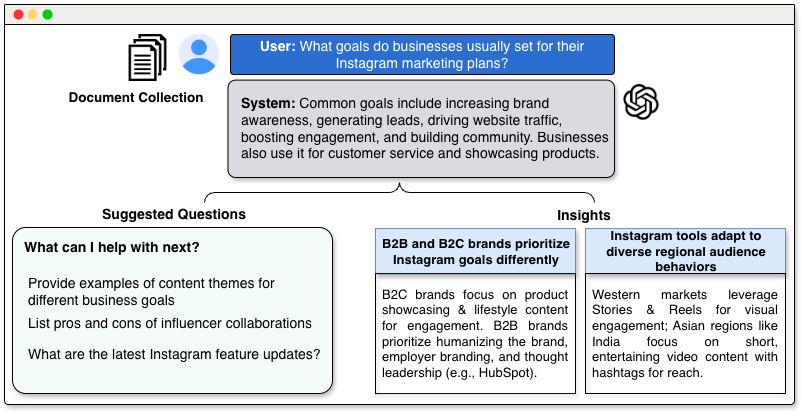}
  \caption{Diagram showing the key difference between insights and traditional follow up or suggested questions. Follow up questions aim to guide the next turn of the conversation, while insights are designed to support the current turn, especially for open ended questions.}
  \label{fig:task_hero}
  \vspace{-5mm}
\end{figure}

Most existing document-grounded QA efforts have primarily focused on answering fact-based questions ({\it e.g., Who is the father of the current president of the United States?}), where there exists a single, objectively correct answer \cite{zhang2025levelragenhancingretrievalaugmentedgeneration,wu2025composeragmodularcomposablerag,zhang2024hierarchicalretrievalaugmentedgenerationmodel}. 
These systems typically rely on retrieving relevant passages, either in a single-hop or multi-hop manner, and generating concise, factual responses.
However, in more real-world scenarios, users interact with their documents with more open-ended queries for various purposes, such as creating reports, formulating strategies, learning, etc. In such settings, it is not straightforward to pinpoint on one “good” answer, as the task demands longer answer statements, more nuanced reasoning processes, and diverse expressions. 
For example, for a question such as {\it Create a draft for a 3-page paper on real world applications of Piaget's theory}, there can be multiple appropriate responses, with different nuances, perspectives, and considerations. 
Similarly, for a question {\it Help me create a 2-page study notes to prepare for an exam}, there can be multiple good responses, say one with a summarized version of the textbook, one with the key insights and common misperceptions, one with a sample quiz and responses, and so on. 

Further, in such settings, a single-shot answer generation framework that generates one final answer, like how most current QA systems are designed, rarely suffices, as users may not be satisfied with the first draft. 
Instead, a framework that enables users to iterate over the generated response by providing AI assistance to brainstorm ideas, refine the content, provide alternate perspectives, or additional data, can more smoothly lead the users to their preferred final version.
Prior works to increase this kind of user engagement largely focused on suggesting follow-up questions or next steps \cite{lin-etal-2025-persona,huang2023readingbots,Wang_Wei_Fan_Liu_Huang_2019}. 
While these offer potential next steps, they also function with the underlying assumption that the generated answer is satisfactory to the user (Figure \ref{fig:task_hero}). 
In this work, we call out the need for open-ended question answering systems to be iterative in nature, and propose that AI assistance should be provided in brainstorming ideas or providing other perspectives to think along for the users, as opposed to being solely focused on improving the quality of the first version of the answers.

In this paper, we introduce the task of \emph{document-grounded related insight recommendation} for open-ended document-grounded question answering, aimed at supporting iterative answer refinement.Given an open-ended question, an initial answer, and a collection of supporting documents, the model is required to generate additional insights that can help improve the existing answer. The generated insights must satisfy three conditions. First, they should be grounded in the provided documents. Second, they should be related to the question while avoiding repetition of information already present in the answer. Third, they should offer useful directions for refining or extending the answer. To get complementary context beyond the current answer, we reimagine K-Means to work on document chunks and build a graph representation of the collection. We then traverse this graph to find information that is related but not repetitive in the current answer. Using a chain-of-thought reasoning prompt~\cite{wei2023chainofthoughtpromptingelicitsreasoning}, this context is processed to generate novel and relevant insights. The graph structure guides the retrieval, enabling a shift from standard similarity to structural complementarity for open-ended insight generation.

This paper makes three main contributions. 
\textbf{(1)} We introduce a new task of document-grounded related insight recommendation for open-ended document QA, and release a new dataset, \textbf{SCOpE-QA} (\textbf{S}cientific \textbf{C}ollections for \textbf{Op}en-\textbf{E}nded \textbf{QA}), consisting of 3,000 open-ended questions spanning 20 research themes. 
\textbf{(2)} We propose a two-stage framework that first retrieves complementary context using a clustering-based neighborhood selection strategy, and then generates insights using a CoT-style reasoning module. 
\textbf{(3)} We compare our approach with naïve prompting and existing RAG-based context selection methods, and show that our method produces insights that are more novel, relevant, and informative.

\begin{figure*}[t]
\centering
  \includegraphics[width=0.85\linewidth]{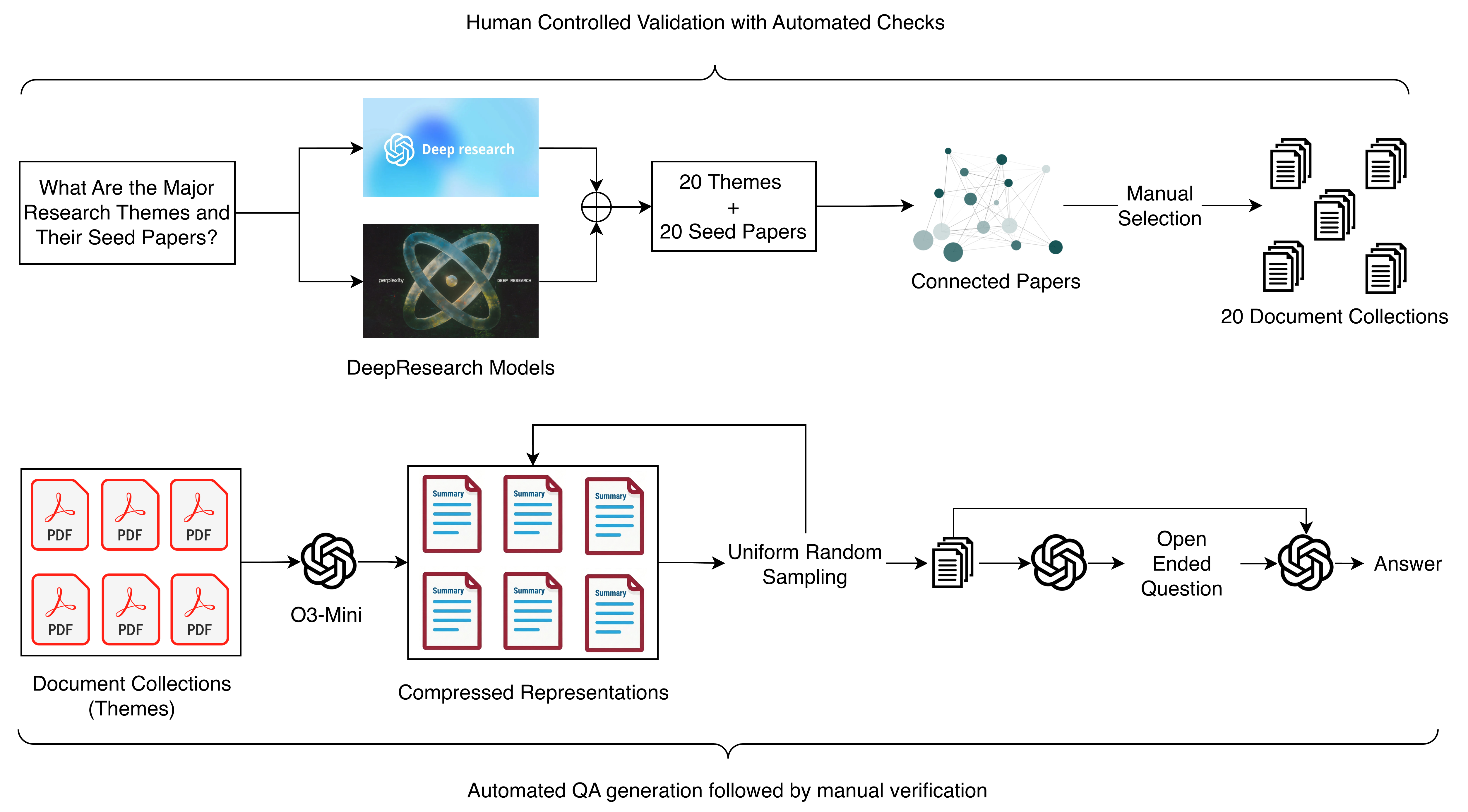}
  \caption{SCOpE-QA dataset curation pipeline overview, highlighting collection curation and QA generation.}
\label{fig:dataset_creation}
  \vspace{-5mm}
\end{figure*}

\section{Related Work and Background}

\textbf{Related Question and Document Understanding:} Prior work most closely related to our task lies in related question generation and document understanding, which are typically applied after an initial answer to guide subsequent interaction. \textbf{Conversational Question Generation (CQG)} focuses on producing follow-up questions in multi-turn settings. Early approaches often ignored previous answers and yielded generic questions \citep{nakanishi-etal-2019-towards, do-etal-2023-modeling}, while later methods improved coherence through answer awareness, reinforcement learning, and discourse modeling \citep{pan-etal-2019-reinforced, gu-etal-2021-chaincqg}, as well as the use of external knowledge and synthetic data \citep{hwang-lee-2022-conversational, faille2024questiongenerationknowledgedrivendialog}. Similarly, \textbf{Suggested Question Generation (SQG)} produces related or clarifying questions, evolving from retrieval and rewriting approaches \citep{10.5555/1625275.1625551} to retrieval-augmented generation with few-shot prompting and personalization \citep{10.1145/3589335.3651905, lin-etal-2025-persona}. Parallel work on document understanding extracts key information through summarization \citep{ijcai2017p574} or structured analysis of research papers \citep{song2025scientific}, although such methods are often constrained by predefined schemas and formats.

While these approaches operate in a similar post-answer setting, they differ fundamentally in objective. Related questions (CQG and SQG) aim to move the conversation forward by identifying what information should be sought next. For example, consider the question: “\textbf{How can a city redesign its public libraries to support digital-era creativity?}” A base answer might suggest maker spaces, digital workshops, startup partnerships, and collaborative work areas. A CQG or SQG system would generate follow-up questions such as “\textbf{What funding models could support these redesigns?}” or “\textbf{How have other cities done this?}”, which extend the discussion by requesting additional information (Figure \ref{fig:task_hero}).

In contrast, insight generation focuses on generating \emph{insights} that analyze and improve the current answer rather than extend it. Insights are statements, not questions, and are designed to surface new perspectives, potential issues, or alternative directions. For the same example, an insight might highlight a concern (“\textbf{Some community members may not like turning libraries into commercial spaces}”), a trade-off (“\textbf{Spending more on digital tools may reduce support for traditional services}”), or a new idea (“\textbf{Libraries could host residency programs for digital artists}”). These statements do not request additional input, but instead help rethink, question, or strengthen the existing answer by identifying missing aspects, risks, or opportunities. In this sense, follow-up questions move the conversation forward, while insights deepen the analysis of what has already been produced.

Our formulation is also distinct from other related tasks such as multi-hop question answering, answer expansion, and multi-document summarization. Multi-hop QA focuses on combining evidence to arrive at an answer, answer expansion adds further details to the same response, and multi-document summarization aggregates and condenses information from multiple sources. In contrast, insight generation operates on the answer itself, examining its assumptions, limitations, and implications. By highlighting trade-offs, surfacing potential issues, and suggesting new directions, insights improve answer quality by helping users reflect on and refine their responses, leading to better outcomes in open-ended tasks such as creative problem solving, critical review, and planning.

\section{SCOpE-QA: Scientific Collections for Open-Ended QA}

Document-grounded insight generation involves three inputs: a collection of related documents, an open-ended question, and its answer. The goal is to produce additional insights grounded in the documents, enabling users to better understand, interact with, or refine the answer. This naturally requires reasoning across multiple documents, reflecting real-world information synthesis. Existing multi-document QA datasets~\citep{zhu2024fanoutqa, bai2024longbench2, wang2024loong} focus on factual retrieval or structured questions and do not support open-ended cross-document reasoning. To address this, we introduce the \textbf{Scientific Collections for Open-Ended QA (SCOpE-QA)} dataset, a publicly released benchmark for multi-document, open-ended reasoning. It contains 20 research collections and 3000 QA pairs, each requiring integration of information across multiple documents. The \texttt{o3-mini} model is used only for dataset construction (document compression and QA generation) and not for baselines or evaluation, making SCOpE-QA a reliable benchmark for studying cross-document insight generation.

\textbf{Collection creation and cleaning:} SCOpE-QA comprises 20 research paper collections across diverse scientific topics. Seed papers and major research themes are identified using \texttt{ChatGPT}\footnote{\url{https://openai.com/index/introducing-deep-research/}} and Perplexity Deep Research\footnote{\url{https://www.perplexity.ai/hub/blog/introducing-perplexity-deep-research}}, which provide representative papers for each theme. Collections are expanded using \texttt{Connected Papers}\footnote{\url{https://www.connectedpapers.com/}}, generating similarity graphs based on co-citations and bibliographic coupling. Papers with high reference overlap are considered related, then filtered automatically via title relevance and abstract similarity, removing non-English, incomplete, or inaccessible documents. Manual review ensures topical relevance based on abstracts and introductions. This process yields 405 documents across 20 collections, each containing 10-35 papers, allowing analysis of context size effects while maintaining high-quality, theme-specific document sets.

\begin{table*}[t]
\centering
\resizebox{0.9\textwidth}{!}{%
\begin{tabular}{lc|rr|rr|rr}
\toprule
\textbf{Research Theme} & 
\textbf{\# Files} & 
\makecell{\textbf{Avg. Tokens}\\\textbf{per File}} & 
\makecell{\textbf{Avg. Words}\\\textbf{per File}} & 
\makecell{\textbf{Avg. Question}\\\textbf{Length (words)}} & 
\makecell{\textbf{Avg. Answer}\\\textbf{Length (words)}} & 
\makecell{\textbf{Avg. Question}\\\textbf{Length (tokens)}} & 
\makecell{\textbf{Avg. Answer}\\\textbf{Length (tokens)}} \\
\midrule
Inference Optimization    & 10 & 22,090.0 & 10,741.1 & 22.3 & 675.0 & 26.9 & 922.8 \\
LLM Agents                & 10 & 23,611.9 & 11,680.3 & 22.1 & 698.1 & 28.4 & 1,005.5 \\
Long-context RAG          & 10 & 23,489.4 & 10,617.1 & 21.3 & 645.7 & 26.5 & 891.6 \\
Preference Optimization   & 10 & 27,719.5 & 16,148.1 & 21.8 & 770.7 & 26.2 & 1,043.3 \\
Hate Speech Detection      & 15 & 20,461.7 & 10,346.1 & 21.6 & 674.9 & 26.2 & 949.3 \\
Long Video Understanding   & 15 & 21,358.1 & 9,968.3  & 21.4 & 641.1 & 25.5 & 884.2 \\
Representation Learning    & 15 & 12,797.0 & 6,741.3  & 22.4 & 643.1 & 28.1 & 896.6 \\
Social Computing           & 15 & 10,538.7 & 6,769.5  & 21.9 & 670.0 & 24.9 & 889.8 \\
Video Segmentation         & 15 & 14,916.3 & 7,370.3  & 21.9 & 591.1 & 27.3 & 810.8 \\
Interpretability           & 20 & 14,119.2 & 7,102.6  & 22.8 & 786.0 & 27.1 & 1,066.1 \\
Low-Resource NLP           & 20 & 21,014.8 & 8,199.6  & 23.7 & 832.0 & 31.4 & 1,211.4 \\
Automatic Speech Recog.    & 25 & 8,319.0  & 4,663.8  & 24.0 & 799.7 & 31.3 & 1,191.1 \\
Data Augmentation          & 25 & 20,189.6 & 9,917.1  & 23.6 & 814.9 & 28.7 & 1,118.8 \\
Ethical Bias \& Fairness   & 25 & 15,579.7 & 8,177.9  & 24.7 & 887.5 & 29.5 & 1,210.6 \\
Legal NLP                  & 25 & 19,057.4 & 9,225.8  & 21.9 & 822.3 & 26.9 & 1,163.8 \\
LLM for Healthcare         & 25 & 28,137.2 & 12,311.9 & 22.2 & 804.9 & 27.5 & 1,156.3 \\
Dialogue Systems           & 30 & 10,908.1 & 6,476.8  & 23.0 & 786.5 & 27.8 & 1,062.1 \\
Quantization               & 30 & 26,699.4 & 10,034.4 & 23.0 & 865.8 & 30.7 & 1,311.3 \\
Reinforcement Learning     & 30 & 14,316.1 & 7,969.5  & 23.5 & 747.4 & 28.3 & 1,022.4 \\
Graph ML                   & 35 & 14,621.5 & 7,810.6  & 24.1 & 777.7 & 29.6 & 1,080.4 \\
\bottomrule
\end{tabular}}
\caption{Summary statistics of the document collection across research themes. The table reports the number of files, average file length in tokens and words, and the average length of questions and answers in both words and tokens.}
\vspace{-6mm}
\label{tab:dataset_stats}
\end{table*}

\textbf{Document compression:} Each document is compressed with \texttt{o3-mini} to retain key information (main topics, conclusions, contributions) while removing non-essential content. This allows multi-document integration when generating QA pairs.

\textbf{Question Generation:} Questions are generated over combinations of compressed papers to encourage cross-document reasoning. For collections with up to 15 papers, 5-paper combinations are sampled, while larger collections use 10-paper combinations. Ten candidate questions are generated per combination, with 50 combinations sampled per collection, yielding about 500 candidates. These are filtered and refined using the \texttt{o3-mini} model to produce 200 high-quality questions per collection (250 for the Quantization).

\textbf{Answer Generation and Verification:} Answers are generated using the same sampled combinations of compressed documents as in question generation, with the model instructed to cite relevant sources. Redundant answers are removed using pairwise cosine similarity with a 0.9 threshold. The remaining QA pairs undergo human verification against the full documents to ensure factual accuracy. After de-duplication and verification, 150 QA pairs are retained per collection (200 for Quantization). For Quantization, these are further split into 150 test and 50 validation pairs.

The final dataset contains 20 collections and 3000 QA pairs. Appendix~\ref{sec:dataset-details} describes the full dataset curation process, and Table~\ref{tab:dataset_stats} summarizes all relevant dataset statistics. Figure~\ref{fig:dataset_creation} summarizes the dataset construction and QA generation pipeline. Apart from the 20 research collections from SCOpE-QA, we also evaluate our pipeline on 15 internal, non-public document collections. These include financial reports, consulting notes, legal documents, and sales materials. Drawn from real organizations, these collections contain real-world questions such as drafting a sales strategy or understanding the business model of a company. We use the same pipeline configuration, prompts, and hyperparameters as in the scientific setting, without any domain-specific changes, so performance on these internal collections reflects the robustness and generalizability of our method.

\section{Problem Formulation and Methodology}
We aim to generate \textit{related insights} as short, reflective prompts that enrich a given QA pair and a document collection. These related insights aim to provide additional perspectives, highlight missing points, and suggest alternative ideas that complement the current answer, helping users think critically and iteratively refine responses. 
We find two main challenges with using existing RAG-based retrieval and naïve prompting strategies to obtain this kind of {\it related context}. First, selecting the right context is difficult: the context should be related to the answer and provide complementary information, but most methods focus only on relevance to the question and ignore adjacency or neighborhood information. 
As we will show in Section \ref{sec:results}, simple relevance-based retrieval strategies underperform in obtaining insightful reflections.
Second, even with suitable context, the generated insights must be \textit{insightful} (deepening the answer), \textit{novel} (providing new information not already in the answer), and \textit{relevant} (focused on the question). 

We hypothesize that in order to generate related insights, we should select context that is related to the given QA pair and not repetitive.
For this, we explicitly represent the document collections using a graph-like representation where the various document chunks are embedded and clustered in semantically similar themes, and in which the adjacency relations capture how similar or different any given themes are.
Given a query, we then identify the most relevant themes to it, and consider them and their neighbors for obtaining appropriate ``related" context for generating the reflections.



\begin{figure*}[t]
\centering
  \includegraphics[width=0.8\linewidth]{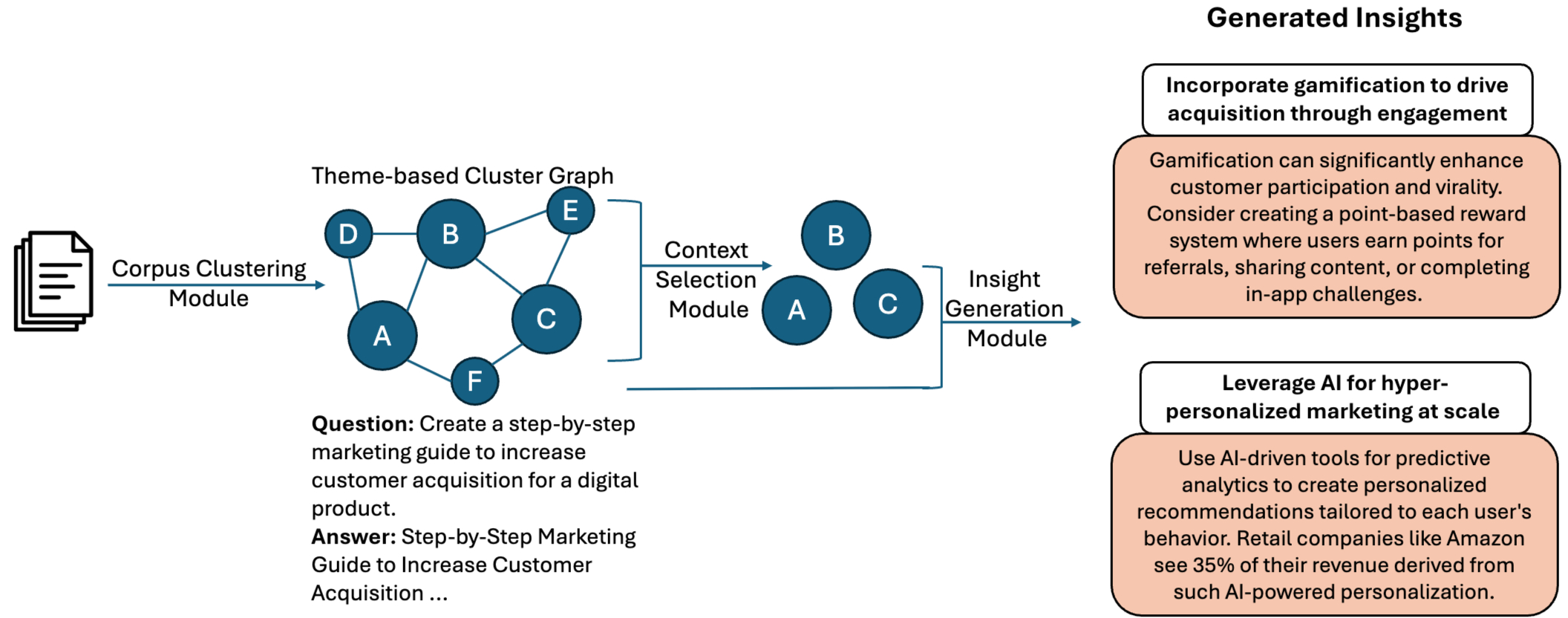}
  \caption{\textsc{InsightGen} pipeline showing theme-based clustering, context selection, and insight generation}
  \label{fig:hero_image}
  \vspace{-6mm}
\end{figure*}
\subsection{Theme-based Document Representations}

To capture the neighborhood relation between different themes or topics, we represent the given document collection using a theme-based graph structure by performing clustering over the document chunks.
For this, we chunk the documents into semantically coherent chunks, preserving sentence boundaries (chunk size of approximately 2K tokens), and embed them using pre-trained Cohere embeddings\footnote{\url{https://cohere.com}}. 
We perform K-means clustering \cite{lloyd1982least} on these embeddings, where each cluster approximately represents a semantic theme or sub-theme in the given collection (we set the number of clusters to \texttt{num\_cluster}). Selecting an appropriate \texttt{num\_cluster} is crucial: too small a value merges loosely related chunks, reducing the diversity of \emph{\textbf{relevant but non-repetitive}} information, while too many clusters lead to overlapping themes and redundant content.

Clustering serves two main purposes: (1) providing a thematic representation of the document collection that clearly identifies key thematic regions, and (2) reducing redundancy by grouping semantically similar chunks. The cluster graph is constructed with edges between centroids representing distances in the embedding (semantic) space, capturing relationships between clusters. Not all clusters are equally relevant to a query, but the \emph{neighborhood surrounding the most relevant clusters} often contains supporting information that is contextually useful and complementary to what the answer itself would cover.

\subsection{Related Insight Generation}
\textbf{Context Selection:}
To identify context relevant to a given answer, we first determine which clusters the answer aligns with most closely. The answer is split into chunks of roughly 2K tokens, preserving semantic boundaries, and each chunk is embedded in the same vector space as the document chunks. For each chunk, we locate the nearest cluster centroids in the cluster graph, and the chunks in these clusters form the \emph{answer-specific context}. To broaden the context, we select the top-$k$ clusters nearest to the answer-specific clusters based on centroid distances; the chunks in these clusters form the \emph{related context}. These clusters contain information \textbf{connected to the main question but not already included in the answer}, providing additional perspectives or supporting details. The hyperparameters $k$ and \emph{max\_hops} (the maximum distance from answer-specific clusters) control the extent of context expansion. Increasing the distance allows access to more \textbf{creative or tangential} information, while careful tuning ensures relevance. 

\textbf{Related Insight Generation:} This module performs the selection of appropriate insight types and the generation of meaningful, actionable insights within a single, systematic process. We define a fixed set of insight types that support learning and creative exploration, including identifying \textit{missing information}, \textit{proposing new ideas}, suggesting \textit{alternate answer} framings, creating \textit{mind maps}, highlighting \textit{potential issues or objections}, presenting \textit{interesting facts}, designing \textit{short quizzes}, providing \textit{real-world applications or analogies}, and analyzing \textit{tradeoffs}. This seed set is designed to aid common personas such as students, analysts, and researchers who routinely work with research paper collections. Given a question–answer pair, the model first infers the user’s intent and persona using a Chain-of-Thought prompt, and then generates insights that add new information beyond the original answer without repeating it, and associates each insight with an appropriate type and a brief justification. The process takes as input the original question, the answer, the selected supporting context, and the inferred user intents and goals, and for each generated insight the model produces self-assessment scores along four dimensions: \textit{relevance}, \textit{novelty}, \textit{usefulness}, and \textit{intent alignment}, each measured on a scale from 0 to 5.

\section{Experimental Setup}
\subsection{Baselines}
Alongside our proposed pipeline, we evaluate four baseline methods to analyze the importance of structured neighborhood selection. \textbf{Direct GPT} uniformly allocates a fixed token budget across all documents and generates insights using a single prompt that specifies requirements such as relevance, novelty, and diversity. \textbf{GPT+CoT} follows the same two-step Chain-of-Thought (CoT) procedure as our pipeline, first inferring user intent and then generating insights, but operates on a fully truncated global context. The CoT prompts are kept identical to avoid any evaluation bias. To isolate the role of clustering-based neighborhood retrieval, we also include two FAISS-based baselines \citep{douze2024faiss}. \textbf{FAISS} retrieves the same number of chunks for each question as those passed to \textsc{\textbf{InsightGen}}, generating insights directly from these chunks, while \textbf{FAISS+CoT} applies the same CoT prompts used in GPT+CoT to the retrieved content. This ensures that any performance difference is due to the quality of the context rather than the amount of context. These baselines help us to verify the hypothesis that relying only on truncated context or similarity-ranked chunks, without explicitly defining local neighborhoods, leads to less diverse and less novel insights.


\begin{table*}[h]
\centering
\small
\begin{adjustbox}{width=\textwidth}
\begin{tabular}{l|l|cc|cc|c|cc|cc|c}
\toprule
\multirow{2}{*}{\textbf{Model}} & \multirow{2}{*}{\textbf{Research Theme}} & \multicolumn{5}{c|}{\textbf{Gemini-2.5-Flash}} & \multicolumn{5}{c}{\textbf{Claude-4-Sonnet}} \\
\cmidrule(lr){3-12}
 & & \textbf{Direct GPT} & \textbf{GPT+CoT} & \textbf{FAISS} & \textbf{FAISS+CoT} & {\sc \textbf{InsightGen}} & \textbf{Direct GPT} & \textbf{GPT+CoT} & \textbf{FAISS} & \textbf{FAISS+CoT} & {\sc \textbf{InsightGen}} \\[1.02ex]
\midrule
\multirow{20}{*}{\textbf{GPT-4o}} 
 & Inference Optimization & \cellcolor{lightred}3.79 & 3.02 & \cellcolor{lightgreen}\textbf{3.87} & 3.03 & \cellcolor{lightgreen}\textbf{3.87} & 2.79 & 1.92 & \cellcolor{lightgreen}\textbf{3.06} & 2.06 & \cellcolor{lightred}3.05 \\[1.02ex]
 & LLM as Agents & 2.99 & 3.25 & \cellcolor{lightred}3.44 & 3.12 & \cellcolor{lightgreen}\textbf{4.14} & 2.04 & 2.32 & \cellcolor{lightred}2.56 & 2.16 & \cellcolor{lightgreen}\textbf{3.37} \\[1.02ex]
 & Preference Optimization & 3.12 & 2.22 & \cellcolor{lightred}3.68 & 3.37 & \cellcolor{lightgreen}\textbf{4.10} & 2.00 & 1.28 & \cellcolor{lightred}2.74 & 2.35 & \cellcolor{lightgreen}\textbf{3.32} \\[1.02ex]
 & Long-context RAG & 3.61 & 3.20 & \cellcolor{lightred}3.81 & 3.25 & \cellcolor{lightgreen}\textbf{4.15} & 2.76 & 2.06 & \cellcolor{lightred}2.86 & 2.30 & \cellcolor{lightgreen}\textbf{3.41} \\[1.02ex]
 & Representation Learning & 3.20 & 2.81 & \cellcolor{lightred}3.55 & 2.93 & \cellcolor{lightgreen}\textbf{4.14} & 2.16 & 1.86 & \cellcolor{lightred}2.63 & 2.05 & \cellcolor{lightgreen}\textbf{3.47} \\[1.02ex]
 & Long Video Understanding & 3.49 & 2.66 & \cellcolor{lightred}3.73 & 3.14 & \cellcolor{lightgreen}\textbf{4.15} & \cellcolor{lightred}2.67 & 1.53 & 2.64 & 2.22 & \cellcolor{lightgreen}\textbf{3.22} \\[1.02ex]
 & Social Computing & 2.82 & \cellcolor{lightred}3.27 & 3.21 & 3.06 & \cellcolor{lightgreen}\textbf{4.52} & 2.16 & 2.26 & \cellcolor{lightred}2.60 & 2.24 & \cellcolor{lightgreen}\textbf{3.84} \\[1.02ex]
 & Video Segmentation & 3.51 & 2.82 & \cellcolor{lightred}3.62 & 2.97 & \cellcolor{lightgreen}\textbf{4.39} & 2.60 & 2.05 & \cellcolor{lightred}2.66 & 2.12 & \cellcolor{lightgreen}\textbf{3.43} \\[1.02ex]
 & Hate Speech Detection & 2.91 & 2.76 & \cellcolor{lightred}3.84 & 3.30 & \cellcolor{lightgreen}\textbf{4.48} & 1.97 & 2.24 & \cellcolor{lightred}2.66 & 2.42 & \cellcolor{lightgreen}\textbf{3.52} \\[1.02ex]
 & Interpretability & 3.24 & 3.13 & \cellcolor{lightred}3.45 & 3.03 & \cellcolor{lightgreen}\textbf{4.23} & 2.24 & 2.21 & \cellcolor{lightred}2.66 & 2.07 & \cellcolor{lightgreen}\textbf{3.57} \\[1.02ex]
 & Low-Resource NLP & 3.35 & 2.80 & \cellcolor{lightred}3.76 & 2.85 & \cellcolor{lightgreen}\textbf{4.34} & 2.20 & 1.82 & \cellcolor{lightred}2.86 & 2.13 & \cellcolor{lightgreen}\textbf{3.60} \\[1.02ex]
 & Data Augmentation & 3.01 & 2.71 & \cellcolor{lightred}3.84 & 2.93 & \cellcolor{lightgreen}\textbf{4.39} & 2.08 & 1.94 & \cellcolor{lightred}3.01 & 2.07 & \cellcolor{lightgreen}\textbf{3.81} \\[1.02ex]
 & Ethical Bias \& Fairness & 3.22 & 2.95 & \cellcolor{lightred}3.33 & 2.85 & \cellcolor{lightgreen}\textbf{4.36} & 2.05 & 1.71 & \cellcolor{lightred}2.49 & 2.43 & \cellcolor{lightgreen}\textbf{3.81} \\[1.02ex]
 & Automatic Speech Recog. & 2.99 & 2.65 & \cellcolor{lightred}3.47 & 2.97 & \cellcolor{lightgreen}\textbf{4.45} & 0.96 & 0.96 & \cellcolor{lightred}2.57 & 2.13 & \cellcolor{lightgreen}\textbf{3.36} \\[1.02ex]
 & LLM for Healthcare & 3.15 & 2.54 & \cellcolor{lightred}3.73 & 3.09 & \cellcolor{lightgreen}\textbf{4.31} & 2.38 & 1.76 & \cellcolor{lightred}2.92 & 2.38 & \cellcolor{lightgreen}\textbf{3.53} \\[1.02ex]
 & Legal NLP & 2.75 & 3.30 & \cellcolor{lightred}3.65 & 3.26 & \cellcolor{lightgreen}\textbf{4.35} & 2.04 & 2.03 & \cellcolor{lightred}2.86 & 2.55 & \cellcolor{lightgreen}\textbf{3.66} \\[1.02ex]
 & Dialogue Systems & 3.06 & 2.71 & \cellcolor{lightred}3.69 & 3.07 & \cellcolor{lightgreen}\textbf{4.27} & 2.39 & 1.91 & \cellcolor{lightred}2.91 & 2.18 & \cellcolor{lightgreen}\textbf{3.53} \\[1.02ex]
 & Reinforcement Learning & 2.88 & 1.90 & \cellcolor{lightred}3.75 & 3.29 & \cellcolor{lightgreen}\textbf{4.27} & 2.08 & 1.14 & \cellcolor{lightred}2.68 & 2.29 & \cellcolor{lightgreen}\textbf{3.56} \\[1.02ex]
 & Quantization & 3.59 & 2.20 & \cellcolor{lightred}3.92 & 3.15 & \cellcolor{lightgreen}\textbf{3.98} & \cellcolor{lightgreen}\textbf{2.96} & 1.24 & 2.78 & 2.09 & \cellcolor{lightred}2.89 \\[1.02ex]
 & Graph ML & 2.71 & 2.11 & \cellcolor{lightred}3.93 & 3.00 & \cellcolor{lightgreen}\textbf{4.26} & 2.06 & 1.65 & \cellcolor{lightred}2.86 & 1.95 & \cellcolor{lightgreen}\textbf{3.63} \\[1.02ex]
\midrule
\multirow{20}{*}{\textbf{Claude-3.5-Sonnet}} 
 & Inference Optimization & 3.18 & 3.32 & 3.09 & \cellcolor{lightred}3.98 & \cellcolor{lightgreen}\textbf{4.45} & \cellcolor{lightred}2.69 & 2.27 & 2.61 & 2.29 & \cellcolor{lightgreen}\textbf{3.11} \\[1.02ex]
 & LLM as Agents & 2.23 & 2.31 & 2.53 & \cellcolor{lightred}4.08 & \cellcolor{lightgreen}\textbf{4.44} & 1.77 & 1.20 & 1.89 & \cellcolor{lightred}2.71 & \cellcolor{lightgreen}\textbf{3.46} \\[1.02ex]
 & Preference Optimization & 2.95 & 2.98 & 2.59 & \cellcolor{lightred}4.07 & \cellcolor{lightgreen}\textbf{4.31} & 2.31 & 2.03 & 2.26 & \cellcolor{lightred}2.63 & \cellcolor{lightgreen}\textbf{3.24} \\[1.02ex]
 & Long-context RAG & 2.29 & 3.54 & 2.81 & \cellcolor{lightred}4.17 & \cellcolor{lightgreen}\textbf{4.41} & 1.67 & 2.56 & 2.21 & \cellcolor{lightred}2.85 & \cellcolor{lightgreen}\textbf{3.33} \\[1.02ex]
 & Representation Learning & 2.37 & 2.50 & 3.13 & \cellcolor{lightred}3.94 & \cellcolor{lightgreen}\textbf{4.33} & 1.68 & 1.27 & \cellcolor{lightred}2.45 & 2.24 & \cellcolor{lightgreen}\textbf{3.27} \\[1.02ex]
 & Long Video Understanding & 2.32 & 3.26 & 3.12 & \cellcolor{lightgreen}\textbf{4.33} & \cellcolor{lightred}4.06 & 1.97 & 2.15 & 2.50 & \cellcolor{lightgreen}\textbf{2.92} & \cellcolor{lightred}2.86 \\[1.02ex]
 & Social Computing & 2.67 & 3.57 & 2.75 & \cellcolor{lightgreen}\textbf{4.07} & \cellcolor{lightred}4.01 & 1.81 & 1.98 & 2.17 & \cellcolor{lightred}2.38 & \cellcolor{lightgreen}\textbf{3.14} \\[1.02ex]
 & Video Segmentation & 2.85 & 2.92 & 3.12 & \cellcolor{lightred}3.68 & \cellcolor{lightgreen}\textbf{4.39} & \cellcolor{lightred}2.64 & 1.95 & 2.35 & 2.02 & \cellcolor{lightgreen}\textbf{3.07} \\[1.02ex]
 & Hate Speech Detection & 1.79 & 2.59 & 2.77 & \cellcolor{lightred}4.24 & \cellcolor{lightgreen}\textbf{4.53} & 1.57 & 1.55 & 2.20 & \cellcolor{lightred}2.85 & \cellcolor{lightgreen}\textbf{3.45} \\[1.02ex]
 & Interpretability & 2.24 & 2.86 & 3.05 & \cellcolor{lightred}4.19 & \cellcolor{lightgreen}\textbf{4.36} & 2.27 & 1.67 & \cellcolor{lightred}2.45 & 2.44 & \cellcolor{lightgreen}\textbf{3.15} \\[1.02ex]
 & Low-Resource NLP & 2.95 & 3.27 & 3.09 & \cellcolor{lightred}4.04 & \cellcolor{lightgreen}\textbf{4.38} & 2.21 & 1.58 & \cellcolor{lightred}2.52 & 2.49 & \cellcolor{lightgreen}\textbf{3.35} \\[1.02ex]
 & Data Augmentation & 3.56 & 3.17 & 3.01 & \cellcolor{lightred}3.79 & \cellcolor{lightgreen}\textbf{3.97} & 2.62 & 2.07 & 2.47 & \cellcolor{lightred}2.57 & \cellcolor{lightgreen}\textbf{3.19} \\[1.02ex]
 & Ethical Bias \& Fairness & 3.17 & 2.37 & 3.19 & \cellcolor{lightred}3.95 & \cellcolor{lightgreen}\textbf{4.34} & 2.33 & 1.77 & \cellcolor{lightred}2.47 & 2.45 & \cellcolor{lightgreen}\textbf{3.41} \\[1.02ex]
 & Automatic Speech Recog. & 3.13 & 2.55 & 3.27 & \cellcolor{lightred}4.08 & \cellcolor{lightgreen}\textbf{4.32} & 2.17 & 1.17 & 2.51 & \cellcolor{lightred}2.55 & \cellcolor{lightgreen}\textbf{3.14} \\[1.02ex]
 & LLM for Healthcare & 2.15 & 2.31 & 2.73 & \cellcolor{lightred}4.19 & \cellcolor{lightgreen}\textbf{4.51} & 1.77 & 1.41 & 2.11 & \cellcolor{lightred}2.80 & \cellcolor{lightgreen}\textbf{3.53} \\[1.02ex]
 & Legal NLP & 2.86 & 3.21 & 2.85 & \cellcolor{lightred}3.90 & \cellcolor{lightgreen}\textbf{4.28} & 2.11 & 2.35 & 2.25 & \cellcolor{lightred}2.45 & \cellcolor{lightgreen}\textbf{3.46} \\[1.02ex]
 & Dialogue Systems & 2.11 & 2.05 & 2.97 & \cellcolor{lightred}3.79 & \cellcolor{lightgreen}\textbf{4.54} & 1.83 & 1.22 & \cellcolor{lightred}2.48 & 2.45 & \cellcolor{lightgreen}\textbf{3.52} \\[1.02ex]
 & Reinforcement Learning & 1.63 & 1.56 & 3.55 & \cellcolor{lightred}3.89 & \cellcolor{lightgreen}\textbf{4.50} & 1.02 & 0.85 & \cellcolor{lightred}2.72 & 2.43 & \cellcolor{lightgreen}\textbf{3.46} \\[1.02ex]
 & Quantization & 2.11 & 2.57 & 3.32 & \cellcolor{lightred}4.03 & \cellcolor{lightgreen}\textbf{4.37} & \cellcolor{lightred}2.61 & 1.99 & 2.44 & 2.40 & \cellcolor{lightgreen}\textbf{3.09} \\[1.02ex]
 & Graph ML & 2.28 & 2.23 & 3.46 & \cellcolor{lightred}3.69 & \cellcolor{lightgreen}\textbf{4.07} & 1.99 & 1.23 & \cellcolor{lightred}2.69 & 2.20 & \cellcolor{lightgreen}\textbf{3.21} \\[1.02ex]
\bottomrule
\end{tabular}
\end{adjustbox}
\caption{Set-level scores across models and themes on research paper collections (SCOpE-QA). Best results are shown in \textcolor{green}{green} and second-best in \textcolor{red}{red}. Insight-level scores are reported in Table~\ref{tab:theme_insight_scores}.
}
\label{tab:theme_set_scores}
\vspace{-5mm}
\end{table*}

\begin{table*}[h]
\centering
\small
\renewcommand{\arraystretch}{1.06} 
\begin{adjustbox}{width=\textwidth}
\begin{tabular}{c|c|ccccc|ccccc}
\toprule
\multirow{2}{*}{\textbf{Model}} & \multirow{2}{*}{\textbf{Domain / Theme}} & \multicolumn{5}{c|}{\textbf{Gemini-2.5-Flash}} & \multicolumn{5}{c}{\textbf{Claude-4-Sonnet}} \\
\cline{3-12}
 & & \textbf{Direct GPT} & \textbf{GPT+CoT} & \textbf{FAISS} & \textbf{FAISS+CoT} & {\sc \textbf{InsightGen}} & \textbf{Direct GPT} & \textbf{GPT+CoT} & \textbf{FAISS} & \textbf{FAISS+CoT} & {\sc \textbf{InsightGen}} \\
\midrule

\multirow{15}{*}{\textbf{GPT-4o}} 
& Legal Business Analysis & 2.94 & 1.12 & 3.59 & \cellcolor{lightred}3.77 & \cellcolor{lightgreen}\textbf{4.65} & 2.00 & 0.77 & 2.77 & \cellcolor{lightred}3.00 & \cellcolor{lightgreen}\textbf{3.47}\\[1.02ex]
& Instagram Marketing & \cellcolor{lightred}3.62 & 3.24 & 3.33 & 3.10 & \cellcolor{lightgreen}\textbf{4.76} & 2.29 & \cellcolor{lightred}2.52 & 2.24 & 2.19 & \cellcolor{lightgreen}\textbf{3.67}\\[1.02ex]
& Climate Change Awareness & \cellcolor{lightred}3.50 & 2.78 & 3.17 & 3.22 & \cellcolor{lightgreen}\textbf{4.78} & \cellcolor{lightred}2.67 & 2.33 & 2.61 & 2.22 & \cellcolor{lightgreen}\textbf{3.78}\\[1.02ex]
& Climate Change Policy & 3.12 & \cellcolor{lightred}3.53 & 2.88 & \cellcolor{lightred}3.53 & \cellcolor{lightgreen}\textbf{4.77} & 2.47 & \cellcolor{lightred}2.82 & 2.41 & 2.71 & \cellcolor{lightgreen}\textbf{2.94}\\[1.02ex]
& Gut Health Insights & 3.55 & 2.70 & \cellcolor{lightred}4.05 & 2.75 & \cellcolor{lightgreen}\textbf{4.15} & 2.85 & 1.95 & \cellcolor{lightred}3.15 & 1.85 & \cellcolor{lightgreen}\textbf{3.45}\\[1.02ex]
& Finance & 3.11 & 3.33 & 3.22 & \cellcolor{lightred}3.67 & \cellcolor{lightgreen}\textbf{4.44} & 2.11 & \cellcolor{lightred}2.67 & 2.11 & 3.00 & \cellcolor{lightgreen}\textbf{3.11}\\[1.02ex]
& Finance - Investment 2 & 2.64 & 3.41 & 3.14 & \cellcolor{lightred}3.46 & \cellcolor{lightgreen}\textbf{4.73} & 2.05 & \cellcolor{lightred}2.41 & 2.14 & 2.23 & \cellcolor{lightgreen}\textbf{3.50}\\[1.02ex]
& Legal \& Regulatory Compliance & 3.40 & 2.80 & \cellcolor{lightred}3.80 & 3.15 & \cellcolor{lightgreen}\textbf{4.25} & 2.50 & 2.10 & \cellcolor{lightred}2.90 & 2.05 & \cellcolor{lightgreen}\textbf{3.00}\\[1.02ex]
& Finance - Investment 3 & 2.73 & \cellcolor{lightred}3.73 & 3.07 & \cellcolor{lightgreen}\textbf{3.80} & \cellcolor{lightred}3.73 & 2.07 & \cellcolor{lightgreen}\textbf{2.67} & 2.20 & \cellcolor{lightred}2.53 & \cellcolor{lightgreen}\textbf{2.67}\\[1.02ex]
& Hotel Sales Strategies & 3.00 & 3.11 & 3.63 & \cellcolor{lightred}3.90 & \cellcolor{lightgreen}\textbf{4.53} & 2.26 & 2.53 & \cellcolor{lightred}2.74 & 2.68 & \cellcolor{lightgreen}\textbf{3.32}\\[1.02ex]
& Responsible AI Consulting & 3.81 & 3.00 & \cellcolor{lightred}3.94 & 2.69 & \cellcolor{lightgreen}\textbf{4.50} & 2.56 & 2.50 & \cellcolor{lightred}3.13 & 2.19 & \cellcolor{lightgreen}\textbf{3.31}\\[1.02ex]
& Revenue \& Finance Reports & 2.70 & 3.25 & \cellcolor{lightred}3.35 & 3.25 & \cellcolor{lightgreen}\textbf{4.45} & 2.15 & 2.45 & 2.55 & \cellcolor{lightred}2.75 & \cellcolor{lightgreen}\textbf{3.25}\\[1.02ex]
& Responsible AI Consulting 2 & 2.70 & \cellcolor{lightred}3.35 & \cellcolor{lightred}3.35 & 3.05 & \cellcolor{lightgreen}\textbf{4.50} & 1.65 & \cellcolor{lightred}2.55 & 2.35 & 2.25 & \cellcolor{lightgreen}\textbf{3.45}\\[1.02ex]
& Summarization of Articles & 2.78 & 3.09 & \cellcolor{lightred}3.78 & 3.26 & \cellcolor{lightgreen}\textbf{4.61} & 2.09 & 2.26 & \cellcolor{lightred}2.70 & 2.35 & \cellcolor{lightgreen}\textbf{3.61}\\[1.02ex]
& Twitter \& Mental Health & 3.13 & 3.25 & \cellcolor{lightred}3.88 & 2.38 & \cellcolor{lightgreen}\textbf{4.31} & 2.31 & 2.25 & \cellcolor{lightred}3.13 & 2.13 & \cellcolor{lightgreen}\textbf{3.63}\\[1.02ex]
\midrule

\multirow{15}{*}{\textbf{Claude-3.5-Sonnet}} 
& Legal Business Analysis & 2.00 & 3.41 & 2.47 & \cellcolor{lightgreen}\textbf{4.65} & \cellcolor{lightred}4.41 & 1.71 & 2.59 & 2.29 & \cellcolor{lightred}2.82 & \cellcolor{lightgreen}\textbf{3.35}\\[1.02ex]
& Instagram Marketing & 3.00 & 3.71 & 1.43 & \cellcolor{lightred}4.24 & \cellcolor{lightgreen}\textbf{4.38} & 1.71 & 2.48 & 1.52 & \cellcolor{lightred}2.62 & \cellcolor{lightgreen}\textbf{3.43}\\[1.02ex]
& Climate Change Awareness & 2.61 & \cellcolor{lightred}4.11 & 1.28 & 3.94 & \cellcolor{lightgreen}\textbf{4.61} & 2.06 & \cellcolor{lightred}3.00 & 1.28 & 2.67 & \cellcolor{lightgreen}\textbf{3.78}\\[1.02ex]
& Climate Change Policy & 2.71 & 3.82 & 1.88 & \cellcolor{lightred}4.12 & \cellcolor{lightgreen}\textbf{4.53} & 1.88 & \cellcolor{lightred}2.82 & 1.65 & \cellcolor{lightgreen}\textbf{2.88} & \cellcolor{lightgreen}\textbf{2.88}\\[1.02ex]
& Gut Health Insights & 3.60 & \cellcolor{lightgreen}\textbf{3.95} & 1.75 & \cellcolor{lightred}3.70 & 3.65 & \cellcolor{lightgreen}\textbf{3.10} & 2.10 & 2.35 & 1.80 & \cellcolor{lightred}2.45\\[1.02ex]
& Finance & 2.44 & \cellcolor{lightred}4.33 & 1.67 & 3.44 & \cellcolor{lightgreen}\textbf{4.56} & 1.78 & \cellcolor{lightred}3.11 & 1.67 & 2.89 & \cellcolor{lightgreen}\textbf{3.33}\\[1.02ex]
& Finance - Investment 2 & 2.46 & 3.82 & 1.82 & \cellcolor{lightred}4.09 & \cellcolor{lightgreen}\textbf{4.41} & 1.68 & \cellcolor{lightred}2.73 & 1.50 & 2.68 & \cellcolor{lightgreen}\textbf{3.41}\\[1.02ex]
& Legal \& Regulatory Compliance & 1.95 & \cellcolor{lightred}4.05 & 2.10 & 3.60 & \cellcolor{lightgreen}\textbf{4.35} & 1.25 & \cellcolor{lightred}2.50 & 1.95 & \cellcolor{lightred}2.50 & \cellcolor{lightgreen}\textbf{2.95}\\[1.02ex]
& Finance - Investment 3 & 2.40 & \cellcolor{lightred}4.20 & 2.13 & 3.87 & \cellcolor{lightgreen}\textbf{4.73} & 1.65 & \cellcolor{lightred}2.82 & 1.82 & 2.59 & \cellcolor{lightgreen}\textbf{3.65}\\[1.02ex]
& Hotel Sales Strategies & 3.21 & 3.63 & 2.37 & \cellcolor{lightred}4.32 & \cellcolor{lightgreen}\textbf{4.53} & 2.00 & 2.42 & 2.05 & \cellcolor{lightred}2.79 & \cellcolor{lightgreen}\textbf{3.21}\\[1.02ex]
& Responsible AI Consulting & 2.63 & 3.25 & 2.31 & \cellcolor{lightred}3.88 & \cellcolor{lightgreen}\textbf{4.31} & 1.44 & 2.13 & 1.94 & \cellcolor{lightred}2.50 & \cellcolor{lightgreen}\textbf{3.50}\\[1.02ex]
& Revenue \& Finance Reports & 1.80 & 3.00 & 2.65 & \cellcolor{lightred}4.25 & \cellcolor{lightgreen}\textbf{4.65} & 1.45 & 2.05 & 2.15 & \cellcolor{lightred}2.95 & \cellcolor{lightgreen}\textbf{3.35}\\[1.02ex]
& Responsible AI Consulting 2 & 2.10 & 1.20 & 2.15 & \cellcolor{lightred}4.25 & \cellcolor{lightgreen}\textbf{4.70} & 1.25 & 0.60 & 1.75 & \cellcolor{lightred}2.80 & \cellcolor{lightgreen}\textbf{3.35}\\[1.02ex]
& Summarization of Articles & 2.96 & 3.48 & 1.87 & \cellcolor{lightred}4.13 & \cellcolor{lightgreen}\textbf{4.39} & 1.91 & 1.87 & 2.04 & 2.22 & \cellcolor{lightgreen}\textbf{3.39}\\[1.02ex]
& Twitter \& Mental Health & 2.63 & 2.88 & 2.69 & \cellcolor{lightgreen}\textbf{4.50} & \cellcolor{lightred}4.38 & 2.19 & 1.88 & \cellcolor{lightred}2.31 & \cellcolor{lightred}2.31 & \cellcolor{lightgreen}\textbf{3.25}\\[1.02ex]
\bottomrule
\end{tabular}
\end{adjustbox}
\caption{Set-level scores across models and domains on internal non-scientific document collections. Best results are shown in \textcolor{green}{green} and second-best in \textcolor{red}{red}. Insight-level scores are reported in Table~\ref{tab:domain_insight_eval}.}
\label{tab:domain_set_eval}
\vspace{-6mm}
\end{table*}

\subsection{Hyperparameter Setup}
\label{sec:hyperparameter-main}
Our pipeline uses three main hyperparameters: \texttt{k}, which determines the number of neighboring clusters selected; \texttt{max\_hops}, which controls how far the traversal proceeds from the answer chunks; and \texttt{num\_cluster}, which denotes the number of clusters used by the KMeans algorithm. Hyperparameter tuning on the validation set of the Quantization collection identifies a single optimal configuration with \texttt{k}=5, \texttt{max\_hops}=2, and \texttt{num\_cluster}=$\sqrt{n}$, where $n$ is the total number of chunks in the collection. The same configuration is applied consistently across all experiments, covering 35 domains and over 3000 questions to validate the robustness and generalizability of the pipeline. Detailed hyperparameter selection and ablation studies are provided in Appendix~\ref{sec:hyperparameter}.

\subsection{Evaluation Metrics}
Each method generates up to five related insights for every QA pair. We use \textbf{GPT-4o} and \textbf{Claude 3.5 Sonnet} as base models for insight generation, and \textbf{Gemini 2.5 Flash} and \textbf{Claude 4 Sonnet} as stronger Judge LLMs for evaluation.


We use two complementary evaluation settings that capture different aspects of insight quality. In \textit{\textbf{Set-Level evaluation}}, the Judge evaluates the complete set of five insights generated by each method and assigns a single score in the range 0-5. The score reflects overall quality based on four criteria: \textbf{Novelty}, which measures how much new information or ideas the insights introduce; \textbf{Diversity}, which captures how distinct the insights are from one another; \textbf{Relevance}, which assesses how well the insights address the original question; and \textbf{Depth}, which indicates whether the insights are substantive rather than superficial. This setting is more robust in that it evaluates the insights as a coherent whole and reflects their overall usefulness. However, as the number of methods or the number of insights per method increases, the evaluation context becomes larger, which can cause the Judge model to under-attend to methods or insights that appear in the middle of the input, potentially affecting score reliability. To address this limitation, we additionally use an \textit{\textbf{Insight-Level evaluation}}, where a single insight is randomly sampled from each method for a given question and evaluated using the same criteria except \textbf{Diversity}. This process is repeated ten times per question and the scores are averaged, making this setting more scalable and less sensitive to long contexts. At the same time, this evaluation incurs higher computational cost as the number of samples increases and may be unreliable if a particularly poor-quality insight is repeatedly sampled. Despite these limitations, the two settings offer complementary perspectives, and using both allows us to evaluate the generated insights from multiple angles.



\section{Results and Discussion}
\label{sec:results}

\textbf{RQ1: How do models perform across diverse domains?}
We evaluate our insight generation approach across diverse domains, including 20 scientific paper collections from \textbf{SCOpE-QA}, (Tables~\ref{tab:theme_set_scores}, \ref{tab:theme_insight_scores}) and 15 non-academic internal collections (Tables~\ref{tab:domain_set_eval}, \ref{tab:domain_insight_eval}). Across all 35 collections, our method consistently matches or outperforms existing baselines, demonstrating the effectiveness of clustering-driven representations and the importance of neighborhood structures for comprehensive insight traversal. In contrast, naively selecting context, whether by taking the first $k$ tokens, the top relevant chunks, or using a chain-of-thought prompt, can result in reduced performance. This highlights that careful context selection is critical for effectively leveraging chain-of-thought reasoning and generating high-quality insights.

\textbf{RQ2: How does collection size affect insight generation?}
In our study of how document collection size affects performance, we observe that for small collections, such as non-academic datasets, Sales, Climate Change, Finance-Investment, Social Computing, or Inference Optimization, full-context methods are often sufficient to generate competitive insights. In these settings, our pipeline maintains strong performance, achieving scores above 4.5 as judged by Gemini. However, as the collection grows larger, including collections like Graph ML, Dialogue Systems, or Quantization, the advantage of clustering-based representations becomes clearly visible. Direct prompting on large collections typically scores between 2-2.5 according to Gemini and even falls below 1 when evaluated by Claude, while our approach consistently achieves scores above 4 with Gemini and above 3 with Claude. This widening gap highlights that clustering-based methods capture broader context and produce higher-quality insights as collection size increases.

\textbf{RQ3: How stable is \textsc{InsightGen} with respect to hyperparameter choices?} We use the hyperparameter configuration described in subsection~\ref{sec:hyperparameter-main}. The same setup generalizes well across all 35 collections, covering more than 3,000 QA pairs, two evaluation models, and two evaluation strategies.To further assess robustness, we conduct a detailed ablation analysis over key hyperparameters, including the clustering method, the number of clusters used in KMeans, the number of neighboring clusters ($k$), and the maximum number of hops. Tables~\ref{tab:rebuttal_num_cluster}, \ref{tab:rebuttal_max_hops}, \ref{tab:rebuttal_clustering}, and \ref{tab:rebuttal_topk} demonstrate that performance remains stable across ablations, confirming the robustness of our selected configuration. Additional details are provided in Appendix~\ref{sec:hyperparameter}.

\textbf{RQ4: How robust are our conclusions to the choice of LLM judge?}
To examine the stability of our conclusions, we compare evaluations from two judges, Claude and Gemini, both of which assign scores on the same 0-5 scale. Although the scoring range is identical, the two evaluators exhibit clearly different calibration behaviors. Claude is substantially stricter, concentrating most of its scores in the lower half of the scale (scores $\leq 3$), resulting in a lower overall mean (approximately 2.4) and very few maximum ratings (around 1\% of scores equal to 5). In contrast, Gemini tends to favor the upper half of the scale (scores $\geq 3$), yielding a higher mean score (approximately 3.3) and assigning the top score much more frequently (about 25\%). Despite these differences in score distributions, both evaluators consistently rank \textsc{InsightGen} as the best-performing method across all settings, indicating that our main conclusions are robust to the choice of LLM judge.

\saransh{\textbf{RQ5: How well does Neighborhood-Aware Retrieval perform compared to advanced Agentic RAG pipelines for insight generation?} We investigate whether insight generation requires fundamentally different retrieval strategies than standard and agentic RAG systems. Prior approaches improve relevance through similarity optimization, query expansion, and iterative refinement, but do not explicitly model structural diversity or inter-document relationships. To evaluate this, we compare \textsc{InsightGen} with three progressively stronger baselines: Iterative RAG, which uses query decomposition and multi-pass self-refinement \cite{asai2023self,chen-etal-2025-sgic}; Multi-Query RAG, which applies query reformulations with LLM-based re-ranking \cite{ma2023query}; and Agentic RAG, which integrates techniques such as HyDE-style embeddings~\cite{gao2023precise}, multi-query expansion \cite{jagerman2023query}, decomposition \cite{ammann-etal-2025-question}, iterative refinement, and re-ranking within a unified framework. For a controlled comparison, all methods use the same hyperparameter and configurations as \textsc{InsightGen}.} Given the rapid evolution of RAG methods, instead of adopting any single end-to-end pipeline as-is, we utilize and combine the core components from these approaches to construct these RAG baselines for our analysis.

\saransh{\textsc{InsightGen} consistently outperforms these baselines (Table: \ref{tab:rag_comparison}), showing that even highly sophisticated, self-refining retrieval pipelines focused on similarity and recall are insufficient for insight generation. In contrast, its neighborhood-aware evidence selection captures complementary structural signals beyond relevance, enabling more effective cross-document synthesis. These results highlight that modeling inter-document structure, rather than solely optimizing retrieval quality, is critical for generating deeper insights.}

\begin{table}[t]
\centering
\small
\begin{adjustbox}{width=0.95\linewidth}
\begin{tabular}{lcccc}
\toprule
\textbf{Theme} & \textbf{InsightGen} & \textbf{Agentic RAG} & \textbf{Iterative} & \textbf{Multi-Query} \\
\midrule
\multicolumn{5}{c}{\textit{Gemini 2.5 Flash}} \\
\midrule
Inference Optimization & \cellcolor{lightgreen}\textbf{4.21} & 3.54 & \cellcolor{lightred}3.74 & 3.44 \\
Dialogue System        & \cellcolor{lightgreen}\textbf{4.14} & 3.50 & \cellcolor{lightred}3.55 & 3.49 \\
Long Video Understanding & \cellcolor{lightgreen}\textbf{4.34} & \cellcolor{lightred}3.73 & 3.44 & 3.60 \\
Social Computing       & \cellcolor{lightgreen}\textbf{4.51} & \cellcolor{lightred}3.52 & 3.27 & 3.13 \\
Video Segmentation     & \cellcolor{lightgreen}\textbf{4.49} & \cellcolor{lightred}3.70 & 3.42 & 3.44 \\
Quantization           & \cellcolor{lightgreen}\textbf{4.09} & 3.57 & \cellcolor{lightred}3.61 & 3.56 \\
\midrule
\multicolumn{5}{c}{\textit{Claude 4 Sonnet}} \\
\midrule
Inference Optimization & \cellcolor{lightgreen}\textbf{2.99} & \cellcolor{lightred}2.58 & 2.48 & 2.56 \\
Dialogue System        & \cellcolor{lightgreen}\textbf{2.75} & \cellcolor{lightred}2.64 & 2.56 & 2.49 \\
Long Video Understanding & \cellcolor{lightgreen}\textbf{2.92} & \cellcolor{lightred}2.67 & 2.62 & 2.52 \\
Social Computing       & \cellcolor{lightgreen}\textbf{3.38} & \cellcolor{lightred}2.62 & 2.51 & 2.37 \\
Video Segmentation     & \cellcolor{lightgreen}\textbf{3.18} & \cellcolor{lightred}2.58 & 2.36 & 2.36 \\
Quantization           & \cellcolor{lightgreen}\textbf{2.67} & 2.51 & \cellcolor{lightred}2.54 & 2.43 \\
\bottomrule
\end{tabular}
\end{adjustbox}
\caption{Comparison with advanced RAG baselines across themes and judge LLMs.}
\label{tab:rag_comparison}
\vspace{-6mm}
\end{table}

\saransh{\textbf{RQ6: How robust is our evaluation across different LLM judges? }Across 3{,}000 questions, cross-judge agreement between Gemini and Claude remains high, with 87\% pairwise ordering agreement, 73\% top-method overlap (top-2 Jaccard: 72\%), and strong rank correlation (median Spearman $\rho = 0.85$; p75 = 0.95; p90 = 1.00; mean within-domain $\rho = 0.83$). All 32 Wilcoxon signed-rank tests comparing \textsc{InsightGen} against baselines remain significant after Bonferroni correction ($\alpha' = 0.00625$), with large effect sizes (median $r = 0.65$), indicating agreement across evaluators.}

\section{Conclusions}
Open-ended question answering over multiple documents requires identifying relevant information across sources while avoiding redundancy, a challenge that standard QA systems often fail to address. To tackle this, we introduce the task of \emph{document-grounded related insight recommendation}, which aims to generate additional insights that help users iteratively refine and explore answers. 
We curate and release \textbf{SCOpE-QA} (\textbf{S}cientific \textbf{C}ollections for \textbf{Op}en-\textbf{E}nded \textbf{QA}), a high-quality dataset covering 20 research paper collections with over 3,000 human-validated question-answer pairs. Building on this dataset, we propose \textsc{InsightGen}, which uses clustering-based context retrieval to generate related insights that promote ideation, brainstorming, highlight overlooked facts, and surface potential limitations in existing answers. Evaluation across 20 research and 15 internal collections, over 3000 QA pairs and 500 documents, using GPT-4o and Claude-3.5 as generation models and Gemini-2.5 and Claude-4 as judges, demonstrates that \textsc{InsightGen} produces more novel, diverse, and higher-quality insights than baseline methods, with benefits increasing as document collections grow larger and more complex.

\section{Limitations and Future Work}
While our results are promising, this work has some limitations that suggest directions for future research. First, the number of generated insights is currently restricted to five due to API constraints. Future work could allow different questions to have a dynamic number of insights based on their complexity and explore strategies to optimize token usage and determine the optimal number of insights per question, ensuring both relevance and coverage while improving scalability and overall quality. Second, our study focuses only on English-language text; extending the approach to multilingual datasets would enable evaluation across different languages and scripts and enhance the generality of the method. Third, the current framework is limited to text data; incorporating multimodal datasets, such as images or structured data, would allow richer insights and better reflect real-world document usage. Addressing these limitations can further improve the flexibility, applicability, and impact of the framework.

\section{Ethical Concerns}
All 20 collections use publicly available, free-to-use papers to avoid copyright issues. Any non-public document details have been appropriately masked to protect copyright and maintain anonymity. Therefore, our work does not raise ethical concerns.

\bibliography{custom}
\appendix

\section{\textbf{S}cientific \textbf{C}ollections for \textbf{Op}en-\textbf{E}nded \textbf{QA} (\textbf{SCOpE-QA}) dataset}

\label{sec:dataset-details}
\subsection{Overview of Collection and Dataset Construction}

Document-grounded related insight generation works with three inputs: a collection of related documents, an open-ended question, and its answer. The goal is to generate additional insights that are grounded in the documents, helping users better understand the question and interact with or refine the answer. This setting naturally requires reasoning across multiple documents, which reflects how people read and combine information in real-world scenarios. However, existing multi-document QA datasets such as FanOutQA \citep{zhu2024fanoutqa}, LongBench v2 \citep{bai2024longbench2}, and Loong \citep{wang2024loong} mainly focus on factual retrieval or structured question formats. They do not capture the open-ended reasoning and cross-document synthesis needed for insight generation. To fill this gap, we introduce the \textbf{S}cientific \textbf{C}ollections for \textbf{Op}en-\textbf{E}nded \textbf{QA} (\textbf{SCOpE-QA}) dataset. SCOpE-QA is a publicly released benchmark designed for open-ended, multi-document reasoning. It includes 20 research themes with collections of varying sizes, and a total of 3000 open-ended question-answer pairs. Each question requires integrating and condensing information spread across multiple documents rather than retrieving isolated facts. The \texttt{o3-mini} model is used only during dataset construction for document compression and question-answer generation, and is not used in insight generation, baselines, or evaluation. These design choices make SCOpE-QA a reliable benchmark for studying document-grounded insight generation, supporting cross-document reasoning and open-ended synthesis of information.

\textbf{1) Collection creation and cleaning:}
The first component of the SCOpE-QA dataset is 20 research paper collections covering a diverse set of scientific topics. To construct these collections, we first identify seed papers and popular research themes using the chat interfaces of \texttt{ChatGPT}\footnote{\url{https://openai.com/index/introducing-deep-research/}} and Perplexity’s Deep Research mode\footnote{\url{https://www.perplexity.ai/hub/blog/introducing-perplexity-deep-research}}. Both tools are prompted to list major research themes along with representative papers for each theme.  Once seed papers are identified, each collection is expanded using \texttt{Connected Papers}\footnote{\url{https://www.connectedpapers.com/}}, which generates a similarity graph based on co-citation patterns and bibliographic coupling. Papers that share a substantial number of references or citations with the seed paper are considered topically related. From the resulting graph, papers are first filtered using automated checks: title relevance and abstract similarity to the seed paper are measured using cosine similarity, and non-English, incomplete, or inaccessible documents are removed. After these automated checks, a careful manual review is performed to validate the topical relevance of each paper based on its abstract and introduction. This combination of automated filtering and manual verification ensures that each collection is focused, coherent, and reliable for downstream analysis. The final dataset contains 405 documents organized into 20 research collections, with each collection containing between 10-35 papers. The variation in collection size allows us to study the effect of context size on model performance, while maintaining high-quality, theme-specific document sets.

\textbf{2) Document Compression:}  
Once the collections are finalized, each document is processed using the \texttt{o3-mini} model to create a compressed, highly informative representation. The compression is performed so that multiple papers can fit within the context when generating question-answer pairs. Since the task focuses on open-ended questions, these compressed representations retain all key information from each paper, including main topics, conclusions, contributions, and essential findings, while removing non-essential material such as formatting markers, reference lists, and boilerplate text. This enables the generation of answers that integrate ideas and contributions spanning multiple documents.

\textbf{3) Question Generation:}  
Using the compressed representations, question generation is performed over combinations of multiple papers to ensure that questions can span ideas across documents. For collections with up to 15 documents, 5-paper combinations are sampled, while for larger collections, 10-paper combinations are used. For each combination, 10 candidate questions are generated. To ensure balanced coverage of the collection, 50 combinations are randomly sampled per collection, resulting in approximately 500 candidate questions. These candidates are then filtered and refined using the \texttt{o3-mini} model to produce 200 high-quality questions per collection (250 for the Quantization theme). 

\textbf{4) Answer Generation and Verification:}  
Answers are generated by prompting the \texttt{o3-mini} model with the same sampled combinations of compressed documents used for question generation. The model is instructed to cite relevant documents whenever a specific source is referenced. To remove highly redundant answers, pairwise cosine similarity is computed between candidate answers, and any answers exceeding a similarity threshold of 0.9 are removed. The remaining question-answer pairs are then subjected to careful human verification for factual correctness. Human annotators consult the full original documents to ensure that answers accurately reflect the content of the papers. After de-duplication and verification, 150 question-answer pairs are retained per collection (200 for the Quantization theme). For Quantization, the QA pairs are further split into 150 for test and 50 for validation set to support hyperparameter tuning.

The final dataset consists of 20 document collections and 3000 QA pairs generated through the multi-step pipeline that combines automated processing with human validation, ensuring high quality for cross-document open-ended question answering. Appendix~\ref{sec:dataset-details} presents an overview of the dataset curation pipeline while Tables~\ref{tab:dataset_stats} and~\ref{tab:domain_dataset_stats} report key statistics, including the number of files, tokens, and words in each collection.

Apart from the 20 research themes from SCOpE-QA, we also evaluate our pipeline on 15 internal, non-public document collections. These include financial reports, consulting notes, legal documents, and sales materials. Drawn from real organizations, these collections contain real-world questions such as, such as drafting a sales strategy or understanding the business model of a company. We use the same pipeline configuration, prompts, and hyperparameters as in the scientific setting, without any domain-specific changes, so performance on these internal collections reflects the robustness and generalizability of our method.

\subsection{Collection Statistics}
\label{sec:corpus_stats}
We provide an overview of the document collections used in our study, summarizing key characteristics across various research themes. Tables~\ref{tab:dataset_stats} and \ref{tab:domain_dataset_stats} present the number of files, average file length in tokens and words, as well as average lengths of questions and answers in both words and tokens. These statistics offer insight into the scale and complexity of each collection, which is relevant for understanding model performance and resource requirements.

\begin{table*}[h]
\centering
\resizebox{\textwidth}{!}{%
\begin{tabular}{lc|cc|cc|cc}
\toprule
\textbf{Domain} & 
\textbf{\# Files} & 
\makecell{\textbf{Avg. Tokens}\\\textbf{per File}} & 
\makecell{\textbf{Avg. Words}\\\textbf{per File}} & 
\makecell{\textbf{Avg. Question}\\\textbf{Length (words)}} & 
\makecell{\textbf{Avg. Answer}\\\textbf{Length (words)}} & 
\makecell{\textbf{Avg. Question}\\\textbf{Length (tokens)}} & 
\makecell{\textbf{Avg. Answer}\\\textbf{Length (tokens)}} \\
\midrule
Legal Business Analysis        & 11 & 15,186.1 & 10,534.4 & 19.9 & 352.1 & 23.5 & 434.1 \\
Instagram Marketing            & 7  & 2,853.0  & 2,169.4  & 18.3 & 363.8 & 20.8 & 457.1 \\
Climate Change Awareness       & 7  & 797.7    & 586.6    & 21.0 & 317.6 & 23.3 & 375.1 \\
Climate Change Policy          & 9  & 12,272.4 & 8,278.3  & 21.7 & 354.1 & 25.8 & 440.9 \\
Gut Health Insights            & 8  & 7,619.1  & 4,113.9  & 24.0 & 377.0 & 29.2 & 519.5 \\
Finance                        & 7  & 6,825.3  & 5,398.6  & 27.0 & 340.1 & 33.3 & 439.0 \\
Finance - Investment 2         & 7  & 6,220.1  & 4,894.0  & 19.6 & 327.3 & 24.0 & 410.0 \\
Legal \& Regulatory Compliance & 8  & 22,948.3 & 16,233.5 & 26.1 & 388.4 & 30.0 & 480.6 \\
Finance - Investment 3         & 9  & 15,812.1 & 5,957.8  & 28.8 & 336.1 & 35.7 & 428.4 \\
Hotel Sales Strategies         & 13 & 2,987.6  & 2,260.1  & 28.8 & 323.0 & 33.8 & 395.5 \\
Responsible AI Consulting      & 13 & 11,631.0 & 8,219.2  & 20.9 & 334.9 & 23.3 & 431.2 \\
Revenue \& Finance Reports     & 13 & 26,116.3 & 15,413.4 & 21.3 & 350.3 & 25.5 & 431.8 \\
Responsible AI Consulting 2    & 13 & 12,180.8 & 7,971.5  & 21.6 & 214.5 & 26.9 & 262.7 \\
Summarization of Articles      & 11 & 10,699.7 & 5,010.6  & 22.4 & 466.9 & 30.0 & 647.4 \\
Twitter \& Mental Health       & 11 & 6,614.7  & 3,788.9  & 25.8 & 382.8 & 29.2 & 495.8 \\
\bottomrule
\end{tabular}}
\caption{Summary statistics of the document collection across domains. The table reports the number of files, average file length in tokens and words, and the average length of questions and answers in both words and tokens.}
\label{tab:domain_dataset_stats}
\end{table*}

\begin{table*}[h]
\centering
\setlength{\tabcolsep}{6pt}
\begin{adjustbox}{width=0.7\textwidth}
\begin{tabular}{l | c c c c | c c c c}
\hline
\multirow{2}{*}{\textbf{Research Theme}} 
& \multicolumn{4}{c|}{\textbf{Gemini 2.5 Flash}} 
& \multicolumn{4}{c}{\textbf{Claude 4 Sonnet}} \\
& $\sqrt{n}$ & $n/3$ & $n/5$ & $\sqrt[3]{n}$ 
& $\sqrt{n}$ & $n/3$ & $n/5$ & $\sqrt[3]{n}$ \\
\hline
Inference Optimization      
& \cellcolor{lightgreen}\textbf{4.28} & 4.11 & 3.52 & 3.28 
& \cellcolor{lightgreen}\textbf{2.98} & 2.94 & 2.55 & 2.14 \\
Dialogue Systems            
& \cellcolor{lightgreen}\textbf{4.04} & 3.58 & 3.89 & 3.48 
& \cellcolor{lightgreen}\textbf{3.05} & 2.69 & 2.67 & 2.50 \\
Long Video Understanding    
& \cellcolor{lightgreen}\textbf{4.37} & 4.02 & 3.72 & 3.13 
& \cellcolor{lightgreen}\textbf{3.03} & 2.77 & 2.52 & 2.17 \\
Social Computing            
& \cellcolor{lightgreen}\textbf{4.37} & 3.68 & 3.63 & 3.48 
& \cellcolor{lightgreen}\textbf{3.29} & 2.88 & 2.70 & 2.51 \\
Video Segmentation          
& \cellcolor{lightgreen}\textbf{4.37} & 3.66 & 3.64 & 3.12 
& \cellcolor{lightgreen}\textbf{3.02} & 2.54 & 2.62 & 2.24 \\
Quantization                
& \cellcolor{lightgreen}\textbf{4.33} & 3.72 & 3.63 & 2.99 
& \cellcolor{lightgreen}\textbf{2.85} & 2.57 & 2.54 & 2.07 \\
\hline
\end{tabular}
\end{adjustbox}
\caption{Ablation on the number of clusters used by KMeans. The hyperparameter \texttt{num\_cluster} controls how many clusters are formed over the chunk set. Results are reported on six research themes selected from \textsc{InsightGen}. Setting the number of clusters to $\sqrt{n}$ gives the best and most consistent performance and is used in the final system.}
\label{tab:rebuttal_num_cluster}
\end{table*}

\begin{table*}[h]
\centering
\small
\setlength{\tabcolsep}{6pt}
\begin{adjustbox}{width=0.7\textwidth}
\begin{tabular}{l | c c c | c c c}
\hline
\multirow{2}{*}{\textbf{Research Theme}} 
& \multicolumn{3}{c|}{\textbf{Gemini 2.5 Flash}} 
& \multicolumn{3}{c}{\textbf{Claude 4 Sonnet}} \\
& 1 Hop & 2 Hop & 3 Hop
& 1 Hop & 2 Hop & 3 Hop \\
\hline
Inference Optimization      
& 3.57 & \cellcolor{lightgreen}\textbf{4.47} & 3.78 
& 2.42 & \cellcolor{lightgreen}\textbf{3.08} & 2.68 \\
Dialogue Systems            
& 3.60 & \cellcolor{lightgreen}\textbf{4.47} & 3.68 
& 2.38 & \cellcolor{lightgreen}\textbf{3.23} & 2.53 \\
Long Video Understanding    
& 3.72 & \cellcolor{lightgreen}\textbf{4.47} & 3.92 
& 2.39 & \cellcolor{lightgreen}\textbf{3.20} & 2.68 \\
Social Computing            
& 3.73 & \cellcolor{lightgreen}\textbf{4.61} & 3.73 
& 2.53 & \cellcolor{lightgreen}\textbf{3.42} & 2.69 \\
Video Segmentation          
& 3.73 & \cellcolor{lightgreen}\textbf{4.53} & 3.82 
& 2.40 & \cellcolor{lightgreen}\textbf{3.23} & 2.60 \\
Quantization                
& 3.28 & \cellcolor{lightgreen}\textbf{4.53} & 3.43 
& 2.18 & \cellcolor{lightgreen}\textbf{2.95} & 2.28 \\
\hline
\end{tabular}
\end{adjustbox}
\caption{Ablation on traversal depth from answer chunks. The hyperparameter \texttt{max\_hops} controls how far retrieval expands from the initial answer chunks. Setting \texttt{max\_hops}=2 gives the best and most consistent performance.}
\label{tab:rebuttal_max_hops}
\end{table*}

\begin{table*}[h]
\centering
\small
\setlength{\tabcolsep}{6pt}
\begin{adjustbox}{width=\textwidth}
\begin{tabular}{l | c c c c | c c c c}
\hline
\multirow{2}{*}{\textbf{Research Theme}} 
& \multicolumn{4}{c|}{\textbf{Gemini 2.5 Flash}} 
& \multicolumn{4}{c}{\textbf{Claude 4 Sonnet}} \\
& KMeans & G-Means & HDBSCAN & X-Means
& KMeans & G-Means & HDBSCAN & X-Means \\
\hline
Inference Optimization      
& \cellcolor{lightgreen}\textbf{4.12} & 3.62 & 3.82 & 3.88 
& \cellcolor{lightgreen}\textbf{2.84} & 2.69 & 2.48 & 2.69 \\
Dialogue Systems            
& \cellcolor{lightgreen}\textbf{4.22} & 3.47 & 3.72 & 3.52 
& \cellcolor{lightgreen}\textbf{3.08} & 2.68 & 2.69 & 2.44 \\
Long Video Understanding    
& \cellcolor{lightgreen}\textbf{4.38} & 3.52 & 3.42 & 3.31 
& \cellcolor{lightgreen}\textbf{3.08} & 2.58 & 2.54 & 2.23 \\
Social Computing            
& \cellcolor{lightgreen}\textbf{4.34} & 3.59 & 3.54 & 3.61 
& \cellcolor{lightgreen}\textbf{3.27} & 2.92 & 2.53 & 2.69 \\
Video Segmentation          
& \cellcolor{lightgreen}\textbf{4.24} & 3.68 & 3.70 & 3.59 
& \cellcolor{lightgreen}\textbf{2.80} & 2.60 & 2.79 & 2.57 \\
Quantization                
& \cellcolor{lightgreen}\textbf{4.32} & 3.37 & 3.08 & 3.40 
& \cellcolor{lightgreen}\textbf{2.83} & 2.37 & 2.23 & 2.33 \\
\hline
\end{tabular}
\end{adjustbox}
\caption{Ablation on the clustering method used in the pipeline. KMeans gives the most stable and best overall performance and is used in the final system.}
\label{tab:rebuttal_clustering}
\end{table*}

\begin{table*}[h]
\centering
\small
\setlength{\tabcolsep}{6pt}
\begin{adjustbox}{width=\textwidth}
\begin{tabular}{l | c c c | c c c}
\hline
\multirow{2}{*}{\textbf{Research Theme}} 
& \multicolumn{3}{c|}{\textbf{Gemini 2.5 Flash}} 
& \multicolumn{3}{c}{\textbf{Claude 4 Sonnet}} \\
& Top-k = 3 & Top-k = 5 & Top-k = 7
& Top-k = 3 & Top-k = 5 & Top-k = 7 \\
\hline
Inference Optimization      
& 3.28 & \cellcolor{lightgreen}\textbf{4.39} & 3.00 
& 2.15 & \cellcolor{lightgreen}\textbf{3.16} & 2.11 \\
Dialogue Systems            
& 3.31 & \cellcolor{lightgreen}\textbf{4.34} & 3.00 
& 2.28 & \cellcolor{lightgreen}\textbf{3.04} & 2.09 \\
Long Video Understanding    
& 2.65 & \cellcolor{lightgreen}\textbf{4.63} & 3.02 
& 1.86 & \cellcolor{lightgreen}\textbf{3.16} & 2.27 \\
Social Computing            
& 3.11 & \cellcolor{lightgreen}\textbf{4.53} & 3.18 
& 2.16 & \cellcolor{lightgreen}\textbf{3.37} & 2.32 \\
Video Segmentation          
& 2.96 & \cellcolor{lightgreen}\textbf{4.40} & 3.28 
& 2.11 & \cellcolor{lightgreen}\textbf{3.07} & 2.38 \\
Quantization                
& 3.06 & \cellcolor{lightgreen}\textbf{4.50} & 3.21 
& 2.12 & \cellcolor{lightgreen}\textbf{2.94} & 2.35 \\
\hline
\end{tabular}
\end{adjustbox}
\caption{Ablation on the number of neighboring clusters retrieved. Setting \texttt{top-k}=5 gives the best and most consistent performance and is used in the final system.}
\label{tab:rebuttal_topk}
\end{table*}

\section{Hyperparameter Setup}
\label{sec:hyperparameter}

Our pipeline uses three core hyperparameters. The parameter \texttt{k} controls the number of neighboring clusters selected during retrieval. The parameter \texttt{max\_hops} determines the maximum traversal depth starting from the answer chunks. The parameter \texttt{num\_cluster} specifies the number of clusters used by the KMeans algorithm. Hyperparameter tuning on the validation split of the Quantization collection yields a single optimal configuration with \texttt{k}=5, \texttt{max\_hops}=2, and \texttt{num\_cluster}=$\sqrt{n}$, where $n$ denotes the total number of chunks in a collection. This configuration is fixed for all experiments and applied uniformly across 35 domains covering more than 3000 questions.

For insight generation, we use GPT-4o and Claude-3.5-Sonnet as the base models. All documents are segmented into chunks of up to 2k tokens using a sentence-boundary-aware chunking strategy. Embeddings for both our pipeline and the FAISS baselines are generated using \texttt{cohere-embed-v3-english}. Clustering is performed using scikit-learn’s KMeans implementation with $\sqrt{n}$ clusters and a fixed random seed of 42 to ensure reproducibility. Evaluation is carried out using two judge LLMs and two evaluation strategies. The consistent performance observed across domains, evaluators, and evaluation protocols indicates that the chosen hyperparameters are stable and not specific to a particular evaluation setup.  

\paragraph{Ablation Setup}
To further validate the robustness of our hyperparameter choices, we conduct ablation studies on six themes, consisting of the top three and bottom three performing themes. All ablations are evaluated using both evaluation models. Unless stated otherwise, the best-performing configuration from validation is used as the default.

\paragraph{Number of Clusters}
We first study the effect of the number of clusters used for KMeans. We evaluate four settings: $\sqrt{n}$, $n/3$, $n/5$, and $\sqrt[3]{n}$, where $n$ denotes the number of chunks in a collection. Table~\ref{tab:rebuttal_num_cluster} shows that the $\sqrt{n}$ configuration consistently performs best across themes and evaluation models. Smaller values tend to merge loosely related chunks, which reduces the diversity of \emph{relevant but non-repetitive} information. In contrast, larger values produce overlapping clusters, leading to redundant context. Overall, $\sqrt{n}$ provides the best balance between coverage and diversity, and is used in all experiments.

\paragraph{Maximum Hops}
Next, we analyze the effect of the maximum traversal distance from answer-specific clusters. We test three values: 1-hop, 2-hop, and 3-hop neighborhood traversal. As shown in Table~\ref{tab:rebuttal_max_hops}, setting \texttt{max\_hops}=2 achieves the best performance. A 1-hop traversal restricts context to very local information, often leading to repetitive insights. Increasing the distance to 3 hops introduces more tangential content, which reduces relevance. The 2-hop setting provides a controlled expansion of context while maintaining relevance, and is used as the default.

\paragraph{Clustering Method}
We also compare different clustering methods, including KMeans, X-Means\cite{pelleg2000x}, G-Means\cite{zhao2008g}, and HDBSCAN\cite{campello2013density}. X-Means and G-Means extend KMeans by automatically adapting the number of clusters based on the data distribution. For X-Means, we initialize the algorithm with KMeans and allow cluster splits based on a model selection criterion, with a fixed random seed of 42 for reproducibility. For G-Means, cluster splitting is guided by a Gaussianity test, using a significance level of 10\% to allow moderate splitting in high-dimensional embeddings. HDBSCAN is a density-based clustering method that does not require specifying the number of clusters. We set the minimum cluster size to 5 and use cosine distance for clustering. Points labeled as noise are reassigned to the nearest cluster centroid to ensure full coverage. Table~\ref{tab:rebuttal_clustering} shows that KMeans provides the most stable performance across themes. Other methods often produce either fewer or more clusters than $\sqrt{n}$, which affects neighborhood structure and downstream retrieval. Using KMeans with $\sqrt{n}$ clusters results in better coverage and a more reliable neighborhood graph.

\paragraph{Number of Neighboring Clusters}
Finally, we study the effect of the number of neighboring clusters selected during retrieval. We evaluate three values for \texttt{top-$k$}. Results in Table~\ref{tab:rebuttal_topk} indicate that selecting 5 neighboring clusters provides the best tradeoff. Smaller values limit context diversity, while larger values introduce redundant or weakly relevant information. Based on this ablation, we set \texttt{top-$k$}=5 for all experiments.

\paragraph{Summary}
These ablation studies demonstrate that our final configuration is robust, despite being tuned on a single validation collection. The same hyperparameter values are applied across all 35 themes without further adjustment, and consistently yield strong performance.

\section{Design and Evaluation Setup}

This section describes the workflow for generating and evaluating domain-specific questions, answers, and insights, along with the key criteria used to assess their quality.

\paragraph{Question Generation} 
We use the o3-mini model with \textit{temperature=0.7, max\_tokens=2000} to generate questions that are exploratory, open-ended, and conceptually rich. Questions are designed to encourage critical thinking and discussion across the domain. For example:

\begin{quote}
``How do different strategies for X reflect underlying assumptions about Y?''
\end{quote}

During generation, questions are constrained to 15–35 words, avoid purely factual or dataset-specific queries, and emphasize conceptual understanding, implications, and trade-offs.

\paragraph{Answer Generation} 
Answers are also produced using o3-mini with \textit{temperature=0.7}. Each answer is structured to be direct, comprehensive, and grounded in the provided collection. Key goals include synthesizing insights from multiple sources, citing supporting evidence with exact paper titles, systematically addressing all aspects of the question, and avoiding information not present in the sources.

\paragraph{Insight Generation} 
Insights provide supplementary perspectives that go beyond the base answer. Each insight consists of a short descriptive hook, a detailed explanation grounded in evidence or examples, and a clear statement of the expected understanding or realization it enables.

We explore two generation settings. In the full context setting, the GPT-4o model (\textit{temperature=0.7, max\_tokens=4000}) receives the entire text collection as input. Generated insights can involve alternative viewpoints, creative brainstorming, knowledge testing, or concept mapping. In the retrieval-based setting, only the top-$k$ most relevant text chunks are used as input via dense retrieval with FAISS and GPT-4o, allowing evaluation of how focused, retrieval-conditioned context affects insight quality and diversity.

\paragraph{Evaluation Setup} 
Insights are evaluated using Gemini 2.5 Flash with scores from 0–5. Evaluation occurs at both the set and individual level. At the set level, four metrics are used: novelty (new ideas beyond the base answer), relevance (connection to the question and user intent), meaningfulness (actionable or informative content), and diversity (coverage of distinct perspectives). At the individual level, diversity is not applied. Each method receives a score along with a brief rationale describing its relative strengths and distinguishing features. This structured evaluation supports both qualitative and quantitative analysis of insight generation performance.

\begin{table*}[h]
\centering
\begin{adjustbox}{width=\textwidth}
\begin{tabular}{c|c|cc|cc|c|cc|cc|c}
\toprule
\multirow{2}{*}{\textbf{Model}} & \multirow{2}{*}{\textbf{Research Theme}} & \multicolumn{5}{c|}{\textbf{Gemini-2.5-Flash}} & \multicolumn{5}{c}{\textbf{Claude-4-Sonnet}} \\
\cmidrule(lr){3-12}
 & & \textbf{Direct GPT} & \textbf{GPT+COT} & \textbf{FAISS} & \textbf{FAISS+COT} & {\sc \textbf{InsightGen}} & \textbf{Direct GPT} & \textbf{GPT+COT} & \textbf{FAISS} & \textbf{FAISS+COT} & {\sc \textbf{InsightGen}} \\[1.02ex]
\midrule
\multirow{20}{*}{\textbf{GPT-4o}}
 & Inference Optimization & \cellcolor{lightgreen}\textbf{4.02} & 3.48 & \cellcolor{lightred}3.98 & 3.42 & \cellcolor{lightred}3.98 & 2.78 & 2.13 & \cellcolor{lightred}2.89 & 2.30 & \cellcolor{lightgreen}\textbf{3.03} \\[1.02ex]
 & LLM as Agents & 3.67 & 3.08 & \cellcolor{lightred}3.73 & 3.56 & \cellcolor{lightgreen}\textbf{4.04} & 2.17 & \cellcolor{lightred}2.54 & 2.53 & 2.34 & \cellcolor{lightgreen}\textbf{3.18} \\[1.02ex]
 & Preference Optimization & 3.37 & \cellcolor{lightred}3.73 & 3.70 & 3.45 & \cellcolor{lightgreen}\textbf{4.07} & 2.33 & 1.81 & \cellcolor{lightred}2.56 & 2.38 & \cellcolor{lightgreen}\textbf{3.13} \\[1.02ex]
 & Long-context RAG & 3.76 & 3.72 & \cellcolor{lightred}3.86 & 3.61 & \cellcolor{lightgreen}\textbf{4.20} & 2.67 & 2.62 & \cellcolor{lightred}2.69 & 2.41 & \cellcolor{lightgreen}\textbf{3.31} \\[1.02ex]
 & Representation Learning & 3.22 & \cellcolor{lightred}3.60 & 3.44 & 3.47 & \cellcolor{lightgreen}\textbf{4.33} & 2.25 & 2.18 & \cellcolor{lightred}2.50 & 2.28 & \cellcolor{lightgreen}\textbf{3.20} \\[1.02ex]
 & Long Video Understanding & 3.37 & 3.39 & \cellcolor{lightred}3.57 & 3.38 & \cellcolor{lightgreen}\textbf{4.08} & 2.65 & 2.10 & \cellcolor{lightred}2.68 & 2.27 & \cellcolor{lightgreen}\textbf{3.24} \\[1.02ex]
 & Social Computing & 3.56 & 3.35 & \cellcolor{lightred}3.71 & 3.32 & \cellcolor{lightgreen}\textbf{4.24} & 2.30 & 2.47 & \cellcolor{lightred}2.50 & 2.35 & \cellcolor{lightgreen}\textbf{3.52} \\[1.02ex]
 & Video Segmentation & 3.20 & 3.52 & \cellcolor{lightred}3.79 & 3.62 & \cellcolor{lightgreen}\textbf{4.31} & 2.56 & 2.29 & \cellcolor{lightred}2.62 & 2.28 & \cellcolor{lightgreen}\textbf{3.27} \\[1.02ex]
 & Hate Speech Detection & 3.70 & 3.25 & \cellcolor{lightred}3.83 & 3.57 & \cellcolor{lightgreen}\textbf{4.20} & 2.18 & 2.41 & \cellcolor{lightred}2.61 & 2.55 & \cellcolor{lightgreen}\textbf{3.39} \\[1.02ex]
 & Interpretability & 3.51 & 3.53 & \cellcolor{lightred}3.61 & 3.26 & \cellcolor{lightgreen}\textbf{4.09} & 2.41 & 2.34 & \cellcolor{lightred}2.57 & 2.30 & \cellcolor{lightgreen}\textbf{3.27} \\[1.02ex]
 & Low-Resource NLP & 3.49 & 3.54 & \cellcolor{lightred}3.86 & 3.40 & \cellcolor{lightgreen}\textbf{4.20} & 2.37 & 2.35 & \cellcolor{lightred}2.74 & 2.31 & \cellcolor{lightgreen}\textbf{3.37} \\[1.02ex]
 & Data Augmentation & 3.51 & \cellcolor{lightred}3.93 & 3.72 & 3.44 & \cellcolor{lightgreen}\textbf{4.24} & 2.32 & 2.20 & \cellcolor{lightred}2.70 & 2.28 & \cellcolor{lightgreen}\textbf{3.39} \\[1.02ex]
 & Ethical Bias \& Fairness & 3.42 & 3.36 & \cellcolor{lightred}3.76 & 3.56 & \cellcolor{lightgreen}\textbf{4.30} & 2.37 & 2.37 & \cellcolor{lightred}2.47 & 2.32 & \cellcolor{lightgreen}\textbf{3.49} \\[1.02ex]
 & Automatic Speech Recog. & 3.49 & 3.32 & \cellcolor{lightred}3.63 & 3.28 & \cellcolor{lightgreen}\textbf{4.19} & 1.39 & 1.31 & \cellcolor{lightred}2.38 & 2.13 & \cellcolor{lightgreen}\textbf{3.21} \\[1.02ex]
 & LLM for Healthcare & 3.57 & \cellcolor{lightred}3.64 & 3.52 & 3.30 & \cellcolor{lightgreen}\textbf{4.15} & 2.43 & 2.13 & \cellcolor{lightred}2.65 & 2.51 & \cellcolor{lightgreen}\textbf{3.40} \\[1.02ex]
 & Legal NLP & 3.30 & 3.27 & \cellcolor{lightred}3.83 & 3.37 & \cellcolor{lightgreen}\textbf{4.21} & 2.34 & 2.70 & \cellcolor{lightred}2.73 & 2.43 & \cellcolor{lightgreen}\textbf{3.43} \\[1.02ex]
 & Dialogue Systems & 3.23 & 2.66 & \cellcolor{lightred}3.70 & 3.27 & \cellcolor{lightgreen}\textbf{4.09} & 2.28 & 2.41 & \cellcolor{lightred}2.63 & 2.30 & \cellcolor{lightgreen}\textbf{3.25} \\[1.02ex]
 & Reinforcement Learning & 3.31 & 3.35 & \cellcolor{lightred}3.72 & 3.34 & \cellcolor{lightgreen}\textbf{4.18} & 2.11 & 1.70 & \cellcolor{lightred}2.51 & 2.27 & \cellcolor{lightgreen}\textbf{3.20} \\[1.02ex]
 & Quantization & 3.80 & 2.99 & \cellcolor{lightred}3.86 & 3.39 & \cellcolor{lightgreen}\textbf{3.89} & \cellcolor{lightred}2.83 & 1.90 & 2.78 & 2.26 & \cellcolor{lightgreen}\textbf{2.99} \\[1.02ex]
 & Graph ML & 3.10 & 3.02 & \cellcolor{lightred}3.74 & 3.33 & \cellcolor{lightgreen}\textbf{4.16} & 2.16 & 1.88 & \cellcolor{lightred}2.62 & 2.31 & \cellcolor{lightgreen}\textbf{3.25} \\[1.02ex]
\midrule
\multirow{20}{*}{\textbf{Claude-3.5-Sonnet}}
 & Inference Optimization & 3.70 & 4.00 & 3.67 & \cellcolor{lightred}4.09 & \cellcolor{lightgreen}\textbf{4.49} & 2.57 & 2.51 & \cellcolor{lightred}2.67 & 2.64 & \cellcolor{lightgreen}\textbf{3.13} \\[1.02ex]
 & LLM as Agents & 3.29 & 3.51 & 3.25 & \cellcolor{lightred}4.05 & \cellcolor{lightgreen}\textbf{4.35} & 2.07 & 1.52 & 2.05 & \cellcolor{lightred}2.71 & \cellcolor{lightgreen}\textbf{3.29} \\[1.02ex]
 & Preference Optimization & 2.92 & 3.01 & 2.94 & \cellcolor{lightred}4.13 & \cellcolor{lightgreen}\textbf{4.47} & 2.32 & 2.25 & 2.32 & \cellcolor{lightred}2.72 & \cellcolor{lightgreen}\textbf{3.15} \\[1.02ex]
 & Long-context RAG & 2.91 & 3.86 & 3.22 & \cellcolor{lightred}4.24 & \cellcolor{lightgreen}\textbf{4.48} & 2.02 & 2.62 & 2.31 & \cellcolor{lightred}2.95 & \cellcolor{lightgreen}\textbf{3.35} \\[1.02ex]
 & Representation Learning & 3.02 & 3.75 & 3.13 & \cellcolor{lightgreen}\textbf{4.12} & \cellcolor{lightred}4.07 & 1.93 & 1.87 & 2.48 & \cellcolor{lightred}2.50 & \cellcolor{lightgreen}\textbf{3.16} \\[1.02ex]
 & Long Video Understanding & 3.15 & 3.21 & 3.64 & \cellcolor{lightred}3.88 & \cellcolor{lightgreen}\textbf{4.37} & 2.09 & 2.39 & 2.42 & \cellcolor{lightred}2.93 & \cellcolor{lightgreen}\textbf{3.10} \\[1.02ex]
 & Social Computing & 3.23 & 3.28 & 3.54 & \cellcolor{lightred}3.85 & \cellcolor{lightgreen}\textbf{4.52} & 2.09 & 2.39 & 2.24 & \cellcolor{lightred}2.76 & \cellcolor{lightgreen}\textbf{2.97} \\[1.02ex]
 & Video Segmentation & 2.43 & 3.03 & 3.15 & \cellcolor{lightred}4.17 & \cellcolor{lightgreen}\textbf{4.47} & 2.38 & 2.13 & 2.44 & \cellcolor{lightred}2.58 & \cellcolor{lightgreen}\textbf{3.24} \\[1.02ex]
 & Hate Speech Detection & 2.94 & 3.57 & 3.51 & \cellcolor{lightred}4.29 & \cellcolor{lightgreen}\textbf{4.33} & 1.74 & 1.99 & 2.18 & \cellcolor{lightred}2.94 & \cellcolor{lightgreen}\textbf{3.33} \\[1.02ex]
 & Interpretability & 3.52 & 3.59 & 3.60 & \cellcolor{lightred}4.03 & \cellcolor{lightgreen}\textbf{4.19} & \cellcolor{lightred}2.60 & 2.24 & 2.48 & 2.51 & \cellcolor{lightgreen}\textbf{2.88} \\[1.02ex]
 & Low-Resource NLP & 3.41 & 3.71 & 3.67 & \cellcolor{lightred}4.13 & \cellcolor{lightgreen}\textbf{4.43} & 2.40 & 2.31 & 2.56 & \cellcolor{lightred}2.73 & \cellcolor{lightgreen}\textbf{3.22} \\[1.02ex]
 & Data Augmentation & 3.45 & 3.87 & 3.35 & \cellcolor{lightred}4.07 & \cellcolor{lightgreen}\textbf{4.39} & 2.50 & 2.46 & 2.62 & \cellcolor{lightred}2.83 & \cellcolor{lightgreen}\textbf{3.11} \\[1.02ex]
 & Ethical Bias \& Fairness & 2.90 & 3.22 & 3.12 & \cellcolor{lightred}4.21 & \cellcolor{lightgreen}\textbf{4.50} & 2.54 & 2.43 & 2.51 & \cellcolor{lightred}2.65 & \cellcolor{lightgreen}\textbf{3.20} \\[1.02ex]
 & Automatic Speech Recog. & 3.45 & 3.40 & 3.62 & \cellcolor{lightred}4.13 & \cellcolor{lightgreen}\textbf{4.42} & 2.21 & 1.89 & 2.44 & \cellcolor{lightred}2.71 & \cellcolor{lightgreen}\textbf{3.10} \\[1.02ex]
 & LLM for Healthcare & 3.60 & 3.67 & 3.51 & \cellcolor{lightred}4.02 & \cellcolor{lightgreen}\textbf{4.32} & 2.00 & 1.84 & 2.17 & \cellcolor{lightred}2.96 & \cellcolor{lightgreen}\textbf{3.44} \\[1.02ex]
 & Legal NLP & 3.61 & 3.75 & 3.64 & \cellcolor{lightred}4.15 & \cellcolor{lightgreen}\textbf{4.27} & 2.40 & 2.67 & 2.32 & \cellcolor{lightred}2.83 & \cellcolor{lightgreen}\textbf{3.33} \\[1.02ex]
 & Dialogue Systems & 2.71 & 2.63 & 3.63 & \cellcolor{lightred}3.80 & \cellcolor{lightgreen}\textbf{4.43} & 2.14 & 1.86 & 2.36 & \cellcolor{lightred}2.60 & \cellcolor{lightgreen}\textbf{3.34} \\[1.02ex]
 & Reinforcement Learning & 2.92 & 2.85 & 3.41 & \cellcolor{lightred}3.82 & \cellcolor{lightgreen}\textbf{4.52} & 1.32 & 1.32 & \cellcolor{lightred}2.54 & 2.49 & \cellcolor{lightgreen}\textbf{3.20} \\[1.02ex]
 & Quantization & 3.08 & 3.33 & 3.72 & \cellcolor{lightred}4.01 & \cellcolor{lightgreen}\textbf{4.42} & 2.48 & 2.30 & 2.56 & \cellcolor{lightred}2.66 & \cellcolor{lightgreen}\textbf{3.18} \\[1.02ex]
 & Graph ML & 2.75 & 2.78 & \cellcolor{lightred}3.70 & 3.68 & \cellcolor{lightgreen}\textbf{4.18} & 1.92 & 1.74 & \cellcolor{lightred}2.55 & 2.34 & \cellcolor{lightgreen}\textbf{2.95} \\[1.02ex]
\bottomrule
\end{tabular}
\end{adjustbox}
\caption{Insight-level scores are reported across models and themes on research paper collections. Best results are highlighted in \textcolor{green}{green}, with second-best results in \textcolor{red}{red}.}
\label{tab:theme_insight_scores}
\vspace{-5mm}
\end{table*}

\begin{table*}[ht]
\centering
\renewcommand{\arraystretch}{1.06} 
\begin{adjustbox}{width=\textwidth}
\begin{tabular}{c|c|ccccc|ccccc}
\toprule
\multirow{2}{*}{\textbf{Model}} & \multirow{2}{*}{\textbf{Domain / Theme}} & \multicolumn{5}{c|}{\textbf{Gemini-2.5-Flash}} & \multicolumn{5}{c}{\textbf{Claude-4-Sonnet}} \\
\cline{3-12}
 & & \textbf{Direct GPT} & \textbf{GPT+COT} & \textbf{FAISS} & \textbf{FAISS+COT} & {\sc \textbf{InsightGen}} & \textbf{Direct GPT} & \textbf{GPT+COT} & \textbf{FAISS} & \textbf{FAISS+COT} & {\sc \textbf{InsightGen}} \\
\midrule

\multirow{15}{*}{\textbf{GPT-4o}} 
& Legal Business Analysis & 3.47 & 1.56 & 3.77 & \cellcolor{lightred}4.12 & \cellcolor{lightgreen}\textbf{4.56} & 2.35 & 0.96 & 2.65 & \cellcolor{lightred}2.82 & \cellcolor{lightgreen}\textbf{3.52}\\[1.02ex]
& Instagram Marketing & \cellcolor{lightred}3.90 & 3.75 & 3.67 & 3.54 & \cellcolor{lightgreen}\textbf{4.59} & \cellcolor{lightred}2.49 & \cellcolor{lightred}2.49 & 2.40 & 2.39 & \cellcolor{lightgreen}\textbf{3.48}\\[1.02ex]
& Climate Change Awareness & \cellcolor{lightred}3.64 & 3.41 & 3.59 & 3.42 & \cellcolor{lightgreen}\textbf{4.64} & \cellcolor{lightred}2.66 & 2.39 & 2.61 & 2.53 & \cellcolor{lightgreen}\textbf{3.53}\\[1.02ex]
& Climate Change Policy & \cellcolor{lightred}3.82 & 3.75 & 3.35 & 3.66 & \cellcolor{lightgreen}\textbf{4.34} & \cellcolor{lightred}2.66 & 2.58 & 2.38 & 2.59 & \cellcolor{lightgreen}\textbf{3.35}\\[1.02ex]
& Gut Health Insights & 3.62 & 3.17 & \cellcolor{lightred}4.01 & 2.97 & \cellcolor{lightgreen}\textbf{4.50} & 2.70 & 2.21 & \cellcolor{lightred}2.97 & 2.03 & \cellcolor{lightgreen}\textbf{3.57}\\[1.02ex]
& Finance & 3.59 & 3.63 & 3.63 & \cellcolor{lightred}3.84 & \cellcolor{lightgreen}\textbf{4.67} & 2.20 & \cellcolor{lightred}2.64 & 2.47 & 2.61 & \cellcolor{lightgreen}\textbf{3.37}\\[1.02ex]
& Finance - Investment 2 & 3.38 & \cellcolor{lightred}3.75 & 3.61 & 3.46 & \cellcolor{lightgreen}\textbf{4.64} & 2.44 & \cellcolor{lightred}2.67 & 2.47 & 2.43 & \cellcolor{lightgreen}\textbf{3.56}\\[1.02ex]
& Legal \& Regulatory Compliance & 3.35 & 3.22 & \cellcolor{lightred}3.63 & 3.60 & \cellcolor{lightgreen}\textbf{4.26} & 2.22 & 2.14 & \cellcolor{lightred}2.63 & 2.53 & \cellcolor{lightgreen}\textbf{3.16}\\[1.02ex]
& Finance - Investment 3 & 3.33 & \cellcolor{lightred}3.96 & 3.47 & 3.83 & \cellcolor{lightgreen}\textbf{4.31} & 2.25 & 2.73 & 2.46 & \cellcolor{lightred}2.77 & \cellcolor{lightgreen}\textbf{3.08}\\[1.02ex]
& Hotel Sales Strategies & 3.44 & 3.52 & \cellcolor{lightred}3.77 & 3.73 & \cellcolor{lightgreen}\textbf{4.52} & 2.33 & 2.50 & 2.54 & \cellcolor{lightred}2.62 & \cellcolor{lightgreen}\textbf{3.42}\\[1.02ex]
& Responsible AI Consulting & 3.76 & 3.56 & \cellcolor{lightred}4.09 & 3.36 & \cellcolor{lightgreen}\textbf{4.64} & 2.56 & 2.50 & \cellcolor{lightred}2.79 & 2.15 & \cellcolor{lightgreen}\textbf{3.48}\\[1.02ex]
& Revenue \& Finance Reports & 3.35 & 3.81 & \cellcolor{lightred}3.92 & 3.84 & \cellcolor{lightgreen}\textbf{4.38} & 2.28 & 2.54 & \cellcolor{lightred}2.66 & 2.65 & \cellcolor{lightgreen}\textbf{3.03}\\[1.02ex]
& Responsible AI Consulting 2 & 3.19 & \cellcolor{lightred}3.92 & 3.42 & 3.69 & \cellcolor{lightgreen}\textbf{4.36} & 2.07 & \cellcolor{lightred}2.53 & 2.38 & 2.19 & \cellcolor{lightgreen}\textbf{3.35}\\[1.02ex]
& Summarization of Articles & 3.38 & 3.53 & \cellcolor{lightred}3.67 & 3.58 & \cellcolor{lightgreen}\textbf{4.38} & 2.40 & 2.34 & \cellcolor{lightred}2.68 & 2.37 & \cellcolor{lightgreen}\textbf{3.27}\\[1.02ex]
& Twitter \& Mental Health & 3.44 & 3.57 & \cellcolor{lightred}3.63 & 3.28 & \cellcolor{lightgreen}\textbf{4.57} & 2.58 & 2.52 & \cellcolor{lightred}2.69 & 2.23 & \cellcolor{lightgreen}\textbf{3.49}\\[1.02ex]
\midrule

\multirow{15}{*}{\textbf{Claude-3.5-Sonnet}} 
& Legal Business Analysis & 3.05 & 4.01 & 2.82 & \cellcolor{lightred}4.60 & \cellcolor{lightgreen}\textbf{4.71} & 1.94 & 2.61 & 1.91 & \cellcolor{lightred}3.18 & \cellcolor{lightgreen}\textbf{3.53}\\[1.02ex]
& Instagram Marketing & 3.19 & 4.06 & 2.02 & \cellcolor{lightred}4.40 & \cellcolor{lightgreen}\textbf{4.76} & 2.25 & 2.68 & 1.50 & \cellcolor{lightred}2.95 & \cellcolor{lightgreen}\textbf{3.37}\\[1.02ex]
& Climate Change Awareness & 3.42 & \cellcolor{lightred}4.16 & 1.47 & 3.92 & \cellcolor{lightgreen}\textbf{4.72} & 2.33 & \cellcolor{lightred}2.81 & 1.16 & 2.69 & \cellcolor{lightgreen}\textbf{3.77}\\[1.02ex]
& Climate Change Policy & 3.18 & 4.15 & 2.63 & \cellcolor{lightgreen}\textbf{4.28} & \cellcolor{lightred}4.19 & 2.11 & \cellcolor{lightred}2.97 & 1.92 & \cellcolor{lightgreen}\textbf{3.07} & 2.60\\[1.02ex]
& Gut Health Insights & 3.62 & \cellcolor{lightgreen}\textbf{4.04} & 2.54 & \cellcolor{lightred}3.88 & \cellcolor{lightgreen}\textbf{4.04} & \cellcolor{lightred}2.72 & 2.48 & 1.81 & 2.53 & \cellcolor{lightgreen}\textbf{2.80}\\[1.02ex]
& Finance & 2.96 & \cellcolor{lightred}4.39 & 1.63 & 3.57 & \cellcolor{lightgreen}\textbf{4.49} & 1.97 & \cellcolor{lightred}3.06 & 1.07 & 2.67 & \cellcolor{lightgreen}\textbf{3.24}\\[1.02ex]
& Finance - Investment 2 & 3.31 & \cellcolor{lightred}4.17 & 2.01 & 3.97 & \cellcolor{lightgreen}\textbf{4.67} & 2.10 & \cellcolor{lightred}2.81 & 1.33 & 2.77 & \cellcolor{lightgreen}\textbf{3.38}\\[1.02ex]
& Legal \& Regulatory Compliance & 2.25 & 3.87 & 2.38 & \cellcolor{lightred}3.88 & \cellcolor{lightgreen}\textbf{4.45} & 1.71 & 2.41 & 1.66 & \cellcolor{lightred}2.59 & \cellcolor{lightgreen}\textbf{3.18}\\[1.02ex]
& Finance - Investment 3 & 2.97 & \cellcolor{lightred}4.35 & 2.84 & 4.07 & \cellcolor{lightgreen}\textbf{4.73} & 1.82 & \cellcolor{lightred}2.88 & 1.72 & 2.52 & \cellcolor{lightgreen}\textbf{3.44}\\[1.02ex]
& Hotel Sales Strategies & 3.56 & 3.94 & 2.29 & \cellcolor{lightred}4.27 & \cellcolor{lightgreen}\textbf{4.36} & 2.35 & 2.62 & 1.63 & \cellcolor{lightred}3.02 & \cellcolor{lightgreen}\textbf{3.32}\\[1.02ex]
& Responsible AI Consulting & 2.70 & 3.88 & 2.69 & \cellcolor{lightred}4.19 & \cellcolor{lightgreen}\textbf{4.56} & 1.75 & 2.58 & 1.88 & \cellcolor{lightred}2.74 & \cellcolor{lightgreen}\textbf{3.36}\\[1.02ex]
& Revenue \& Finance Reports & 2.66 & 3.54 & 3.10 & \cellcolor{lightred}4.40 & \cellcolor{lightgreen}\textbf{4.62} & 1.52 & 2.16 & 2.04 & \cellcolor{lightred}3.02 & \cellcolor{lightgreen}\textbf{3.38}\\[1.02ex]
& Responsible AI Consulting 2 & 2.97 & 2.46 & 2.40 & \cellcolor{lightred}4.08 & \cellcolor{lightgreen}\textbf{4.57} & 1.46 & 0.81 & 1.47 & \cellcolor{lightred}2.65 & \cellcolor{lightgreen}\textbf{3.26}\\[1.02ex]
& Summarization of Articles & 3.34 & 3.67 & 2.39 & \cellcolor{lightred}3.68 & \cellcolor{lightgreen}\textbf{4.46} & 2.38 & 2.37 & 1.70 & \cellcolor{lightred}2.45 & \cellcolor{lightgreen}\textbf{3.11}\\[1.02ex]
& Twitter \& Mental Health & 2.99 & 3.34 & 3.16 & \cellcolor{lightred}3.98 & \cellcolor{lightgreen}\textbf{4.46} & 2.24 & 2.18 & 2.46 & \cellcolor{lightred}2.56 & \cellcolor{lightgreen}\textbf{3.36}\\[1.02ex]
\bottomrule
\end{tabular}
\end{adjustbox}
\caption{Insight-level scores across models and domainson internal document collections (non-scientific domain). Best results are highlighted in \textcolor{green}{green}, with second-best results in \textcolor{red}{red}.}
\label{tab:domain_insight_eval}
\vspace{-5mm}
\end{table*}

\section{Insight-level Evaluation}
Different methods tend to generate different types of insights for the same question. The underlying type of insight can also vary depending on how much context is visible during insight generation. Because of this design choice, we adopt a \textbf{set-level evaluation} strategy. In this setting, the LLM judge evaluates the generations from all five methods as a single set. This helps avoid over-penalizing one poor insight or over-rewarding a single strong insight within the set.

However, when the number of insights varies across methods, or when the number of methods or insights increases, context length can become a limitation. To test the robustness of set-level evaluation, we also propose an \textbf{insight-level evaluation}. For each question, one insight is randomly sampled from each method and evaluated using the same criteria, except for Diversity. This process is repeated ten times per question, and the final score is computed by averaging across runs.

Since insights are highly subjective, both evaluation settings use comparative prompts. The order of methods is randomly shuffled for every question to reduce positional bias. Tables \ref{tab:theme_insight_scores} and \ref{tab:domain_insight_eval} summarize the key findings. \textsc{InsightGen} consistently outperforms the baselines on the 20 research paper collections and remains competitive on the 15 internal non-scientific collections. This demonstrates the robustness of our method across evaluation strategies. The insight-level evaluation serves as a secondary confirmation of the set-level results, showing that our method remains stronger even when evaluation is performed at the level of individual insights.

\section{Thematic Summaries of Collections}
To highlight the domains, sub-domains, and topics shared across the 35 collections (20 public research collections and 15 internal collections), we provide thematic summaries to better understand the focus of the papers.

\subsection{Research Paper Collections: Fully Public}

\begin{itemize}

\item \textbf{Inference Optimization:} This collection focuses on optimizing inference for large neural networks, mainly LLMs and Transformer-based models. It covers techniques like early exiting, adaptive layer skipping, block skipping, and parallel decoding. The main goal is to reduce inference latency and computational cost while maintaining model performance.

\item \textbf{LLM Agents:} This collection explores LLM Agents, emphasizing reasoning, acting, and planning. It includes reinforcement learning, self-reflection, and tool use, aiming to build robust frameworks and architectures to improve the autonomy and capabilities of language model agents.

\item \textbf{Preference Optimization:} This collection focuses on LLM alignment, exploring preference optimization methods like RLHF and Direct Preference Optimization. It covers algorithms and theoretical frameworks aimed at improving LLM behavior to match human preferences for helpfulness and harmlessness.

\item \textbf{Long-context RAG:} This collection studies Long-context Retrieval-Augmented Generation (RAG) and LLMs, focusing on long-context understanding, retrieval vs extended context comparison, and novel architectures for handling long inputs efficiently.

\item \textbf{Representation Learning:} This collection covers representation learning in NLP, including vector representations for words, phrases, and sentences. It explores neural architectures and training paradigms to develop effective and transferable representations.

\item \textbf{Long Video Understanding:} This collection focuses on understanding long videos with multimodal LLMs, addressing context length, computation, and redundancy. Techniques include token compression, frame selection, and temporal reasoning for hour-long videos.

\item \textbf{Social Computing:} This collection studies user characteristics and dynamics on social platforms like Twitter. Topics include measuring influence, identifying experts, and understanding interests, leveraging content and social network structure.

\item \textbf{Video Segmentation:} This collection investigates Video Object Segmentation, focusing on unsupervised and zero-shot approaches. It explores architectures and methods that combine appearance, motion, and other modalities for robust automatic segmentation.

\item \textbf{Hate Speech Detection:} This collection studies detecting and explaining hateful content in multimodal memes. Approaches include classification, target identification, and generating rationales, often using language and vision-language models.

\item \textbf{Interpretability:} This collection explores the interpretability of deep neural networks, mainly LLMs like Transformers. It focuses on understanding linguistic knowledge encoded in attention mechanisms and contextual representations.

\item \textbf{Low-Resource NLP:} This collection studies low-resource and multilingual NLP, including cross-lingual transfer, embedding alignment, vocabulary generation, and architecture design to improve performance across diverse languages.

\item \textbf{Data Augmentation:} This collection covers data augmentation in NLP, including rule-based, contextual, and generative methods applied to tasks like text classification and sequence labeling, aiming to address data scarcity and improve model performance.

\item \textbf{Ethical Bias \& Fairness:} This collection investigates bias and fairness in AI, focusing on NLP and computer vision. It studies detection, quantification, and mitigation of biases like gender, racial, and intersectional stereotypes in embeddings, models, and classifiers.

\item \textbf{Automatic Speech Recognition:} This collection focuses on improving end-to-end ASR models, including contextual biasing, personalization, and handling rare or domain-specific words. It explores integrating external knowledge, language models, and phonetic information.

\item \textbf{LLM for Healthcare:} This collection advances large multimodal models for healthcare, mainly in medical vision-language tasks. Topics include foundation models, domain-specific datasets, visual question answering, report generation, and diagnosis.

\item \textbf{Legal NLP:} This collection studies NLP for legal documents, covering automatic summarization of cases, judgments, and verdicts, as well as prediction and explanation of legal outcomes, across languages and jurisdictions.

\item \textbf{Dialogue Systems:} This collection explores advanced ML approaches for dialogue systems, mainly task-oriented dialogue. It emphasizes reinforcement learning for dialogue policy optimization, exploration, sample efficiency, and robustness.

\item \textbf{Reinforcement Learning:} This collection focuses on reinforcement learning, exploring policy optimization techniques such as trust region methods, maximum entropy, and deterministic policy gradients for continuous control, with emphasis on stability and sample efficiency.

\item \textbf{Quantization:} This collection studies quantization for neural networks, mainly LLMs, covering post-training and quantization-aware methods for weights, activations, and KV caches to improve efficiency on resource-limited devices.

\item \textbf{Graph ML:} This collection focuses on graph machine learning and Graph Neural Networks, covering representation learning, semi-supervised learning, and architectural innovations like graph convolutions, attention, and hierarchical pooling for complex graph-structured data.

\end{itemize}

\begin{figure*}[h]
  \includegraphics[width=0.48\linewidth, height=5.5cm]{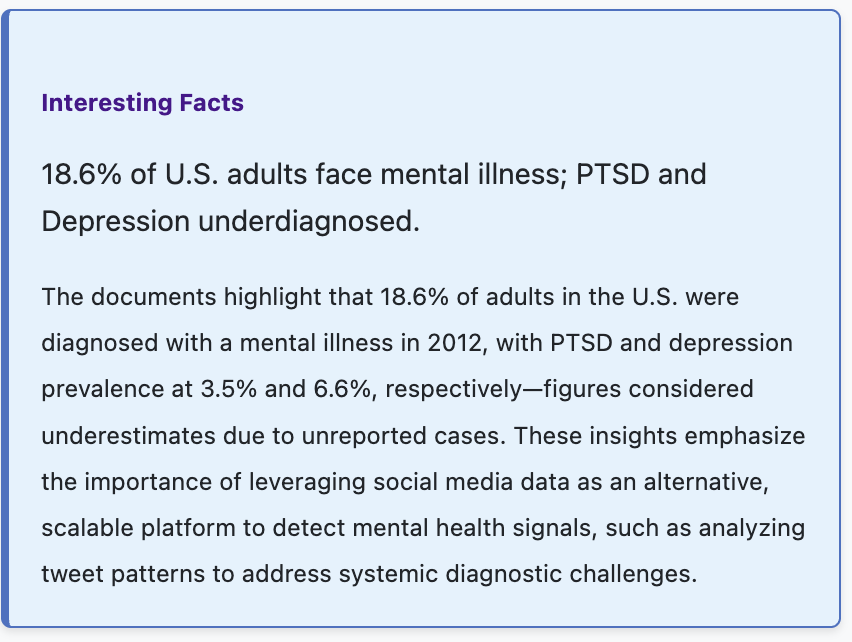}\hspace{0.02\linewidth}
  \includegraphics[width=0.48\linewidth, height=5.5cm]{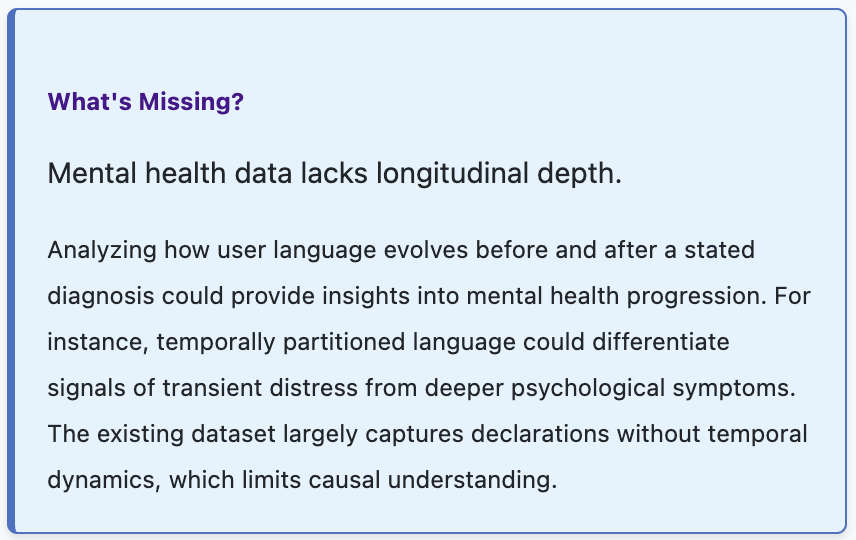}
  \caption{Example insights showing differences in type and content. These insights provide additional observations and reflections from a set of documents on analyzing mental health on Twitter.}
  \label{fig:sample_insights}
  \vspace{-3mm}
\end{figure*}

\begin{figure*}[h]
  \includegraphics[width=0.48\linewidth, height=5.5cm]{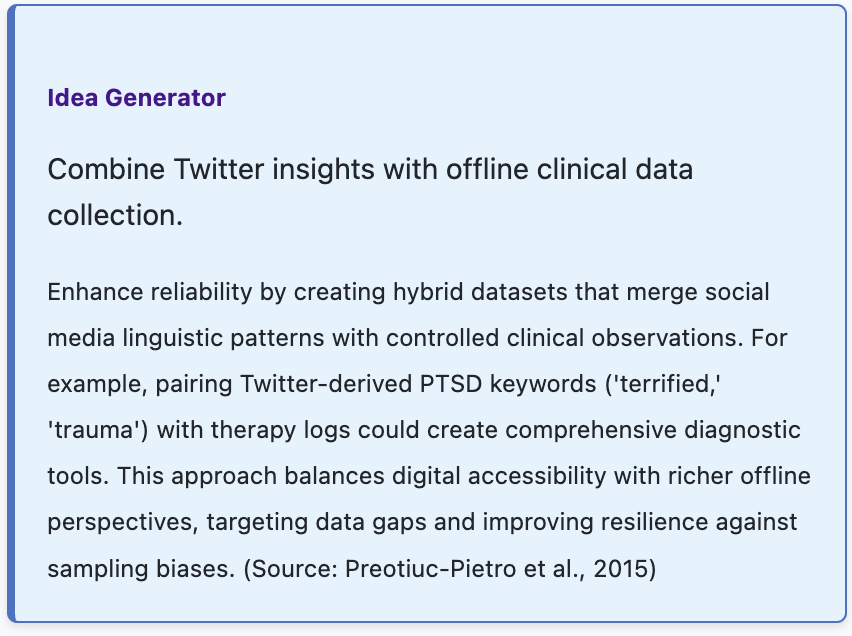}\hspace{0.02\linewidth}
  \includegraphics[width=0.48\linewidth, height=5.5cm]{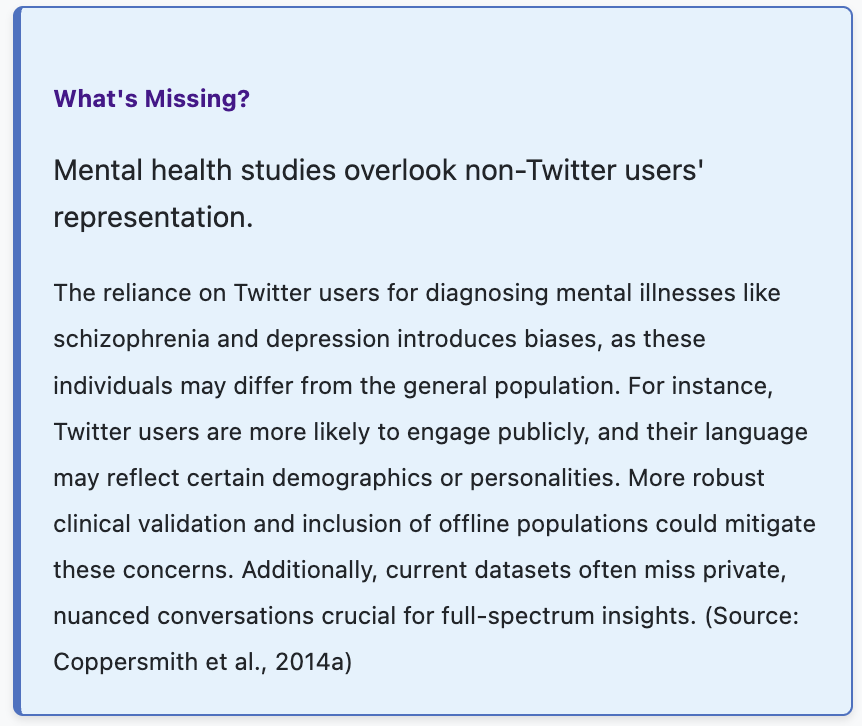}
  \caption{ Example insights showing differences in type and content. These insights provide additional observations and reflections from a set of documents on analyzing mental health on Twitter.}
  \label{fig:sample_insights-2}
  \vspace{-2mm}
\end{figure*}

\section{Qualitative Examples}

\subsection{Example insights generated for different questions}
Figures~\ref{fig:sample_insights} and \ref{fig:sample_insights-2} show what we mean by insights. For an open-ended question like “What are the key findings of these documents?” on a collection discussing mental health analysis on Twitter, these insights go beyond simply summarizing the text. They draw attention to interesting facts, suggest new ideas, or highlight actionable points that can be included to improve answers. In this way, they become useful tools to enhance user interaction, make existing answers better, and improve overall user experience and engagement.

Table~\ref{tab:method_comparison} presents selected insights generated by different baselines. To keep the table clear, we show only the best-performing variant for each method, with or without CoT. The examples clearly illustrate the superior quality of \textsc{InsightGen} and show that Gemini’s judgments closely match manual inspection.

\begin{table*}[t]
\centering
\scalebox{0.7}{
\begin{tabular}{p{4cm}|p{1.5cm}|p{2cm}|p{13cm}}
\toprule
\textbf{Question} & \textbf{Domain} & \textbf{Method} & \textbf{Response} \\
\midrule

\multirow{4}{=}{Give me campaign ideas for a product launch of a beauty brand based on these guidelines} 
& \multirow{4}{=}{Marketing} 
& \textsc{InsightGen} 
& Inspire users with a 7-day transformation challenge. Engage audiences via a \#7DaysToGlow campaign using daily Stories and branded hashtags, encouraging UGC and giveaways for the best transformations. \vspace{0.5em} \\
& & GPT
& Instagram is a critical platform for beauty brands due to its visual engagement. 81\% use it to inform purchases, making it vital for visually-driven products. \vspace{0.5em} \\
& & FAISS 
& Leverage Instagram’s power where 81\% of users research products. Visual content is key to driving conversions and business site visits. \vspace{0.5em} \\
& & Gemini Judgement 
& Method 1 offers novel, actionable, and diverse insights, while Method 2 repeats key points and lacks consistent novelty, and Method 3 mostly rephrases existing ideas with limited depth or originality. \vspace{0.5em} \\
\midrule

\multirow{4}{=}{Can you revise "The \textless HOTEL-CHAIN\textgreater Effect" paper to make it more engaging and convincing?} 
& \multirow{4}{=}{Sales} 
& \textsc{InsightGen} 
& The paper overlooks recent travel trends shaping consumer behaviors. Consider integrating traveler archetypes (e.g., culture seekers, sun-and-beach lovers) and data showing 18\% prioritize authentic experiences, using \textless HOTEL-CHAIN\textgreater's regional cuisine programs and local artisan tie-ups as examples. \vspace{0.5em} \\
& & GPT
& Highlight \textless HOTEL-CHAIN\textgreater’s milestones to underscore its transformative role in the travel industry. Mention TV installations (1947), room service (1930), and global reservation systems (1947) to reinforce its legacy. \vspace{0.5em} \\
& & FAISS
& Incorporate facts about \textless HOTEL-CHAIN\textgreater’s innovations, TVs in rooms, reservation hotlines, and Connected Room concepts, to capture attention, though these are mostly restatements from source chunks. \vspace{0.5em} \\
& & Gemini Judgement 
& Method 1 consistently adds novel, strategic insights that go beyond the original answer, unlike Method 2’s moderate expansions and Method 3’s occasional repetition. \vspace{0.5em} \\
\midrule

\multirow{4}{=}{How has the integration of contrastive reasoning, retrospective optimization, and episodic memory evolved in frameworks aimed at enhancing LLM as Agents' decision-making capabilities?} 
& \multirow{4}{=}{Scientific} 
& \textsc{InsightGen} 
& Explore how contrastive reasoning, optimization, and memory can enhance real-world autonomous systems, e.g., dynamic routing in AVs, action refinement in robotics, and personalized learning based on historical interactions, as shown in Reflexion’s cross-domain gains in robustness and success rates. \vspace{0.5em} \\
& & GPT
& Episodic memory enriches retrospective optimization by providing contextual feedback, enabling dynamic learning from failure. Reflexion leverages this for self-reflective learning, which could be enhanced with policy gradients for iterative policy refinement. \vspace{0.5em} \\
& & FAISS
& Episodic memory feeds retrospective optimization with contextual feedback, enabling adaptive failure-driven learning. Reflexion demonstrates this, with potential gains from adding policy gradients for iterative policy refinement. \vspace{0.5em} \\
& & Gemini Judgement 
& Method 1 consistently delivers novel, deep, and diverse insights with clear examples, while Method 2 lacks novelty, and Method 3 suffers from rule violations. \vspace{0.5em} \\
\bottomrule
\end{tabular}}
\caption{Comparison of insights across domains and methods. Method 1 is our pipeline, Method 2 uses direct GPT-4o prompting, and Method 3 combines FAISS-based retrieval with GPT-4o. Our pipeline consistently generates novel and actionable insights, while other methods tend to rephrase content. For Methods 2 and 3, we show the variant (with or without chain-of-thought prompts) that achieves the higher score for a clearer comparison.}
\label{tab:method_comparison}
\end{table*}

\subsection{Collection Visualization and Thematic Clustering}
\begin{figure*}[h]
\centering
\includegraphics[width=\linewidth]{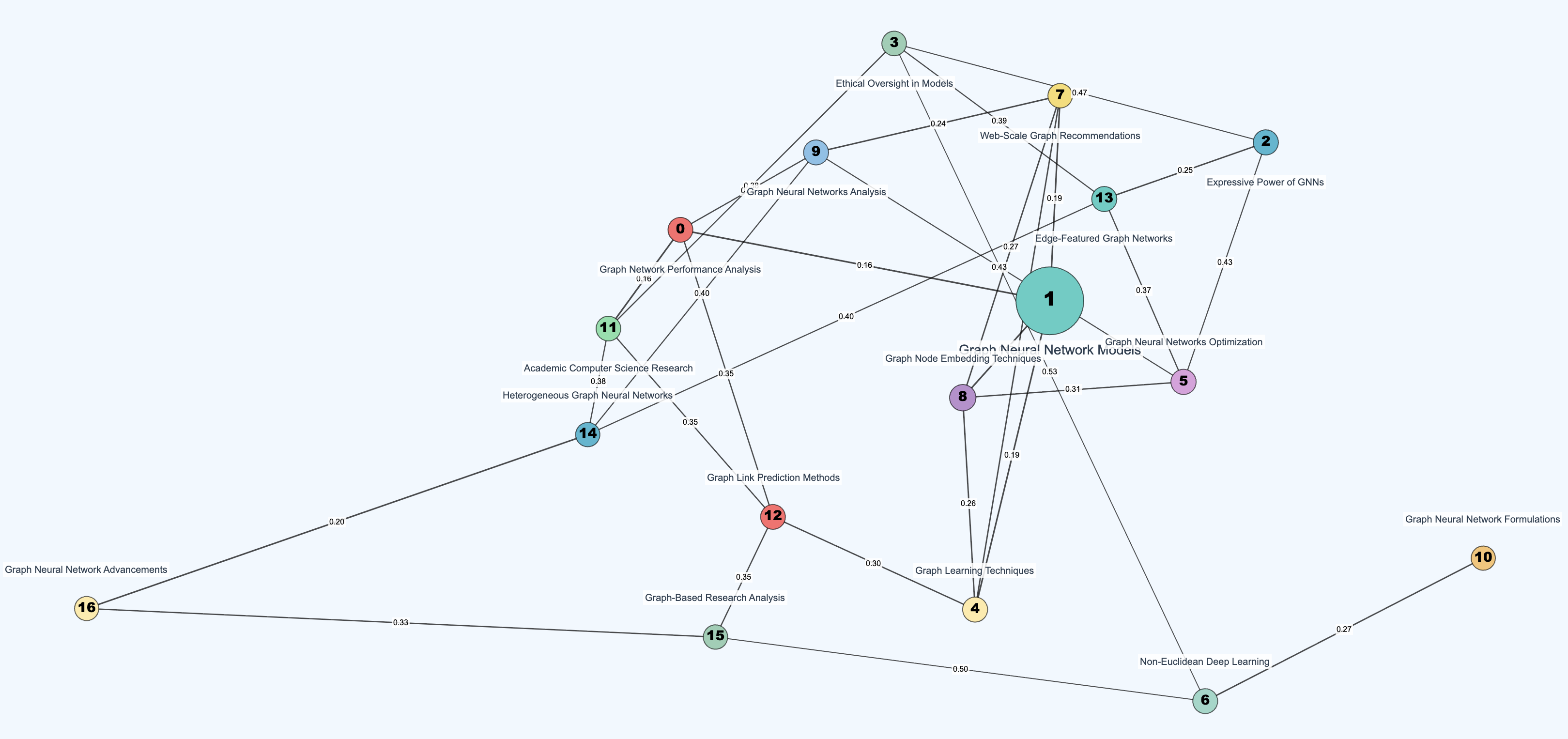}
\caption{Thematic clusters from the Graph ML collection. Each node corresponds to a document segment, and edges represent semantic relationships. Clustering highlights distinct research themes within the Graph ML domain.}
\label{fig:hate_speech_clusters}
\vspace{-3mm}
\end{figure*}

\begin{figure*}[h]
\centering
\includegraphics[width=\linewidth]{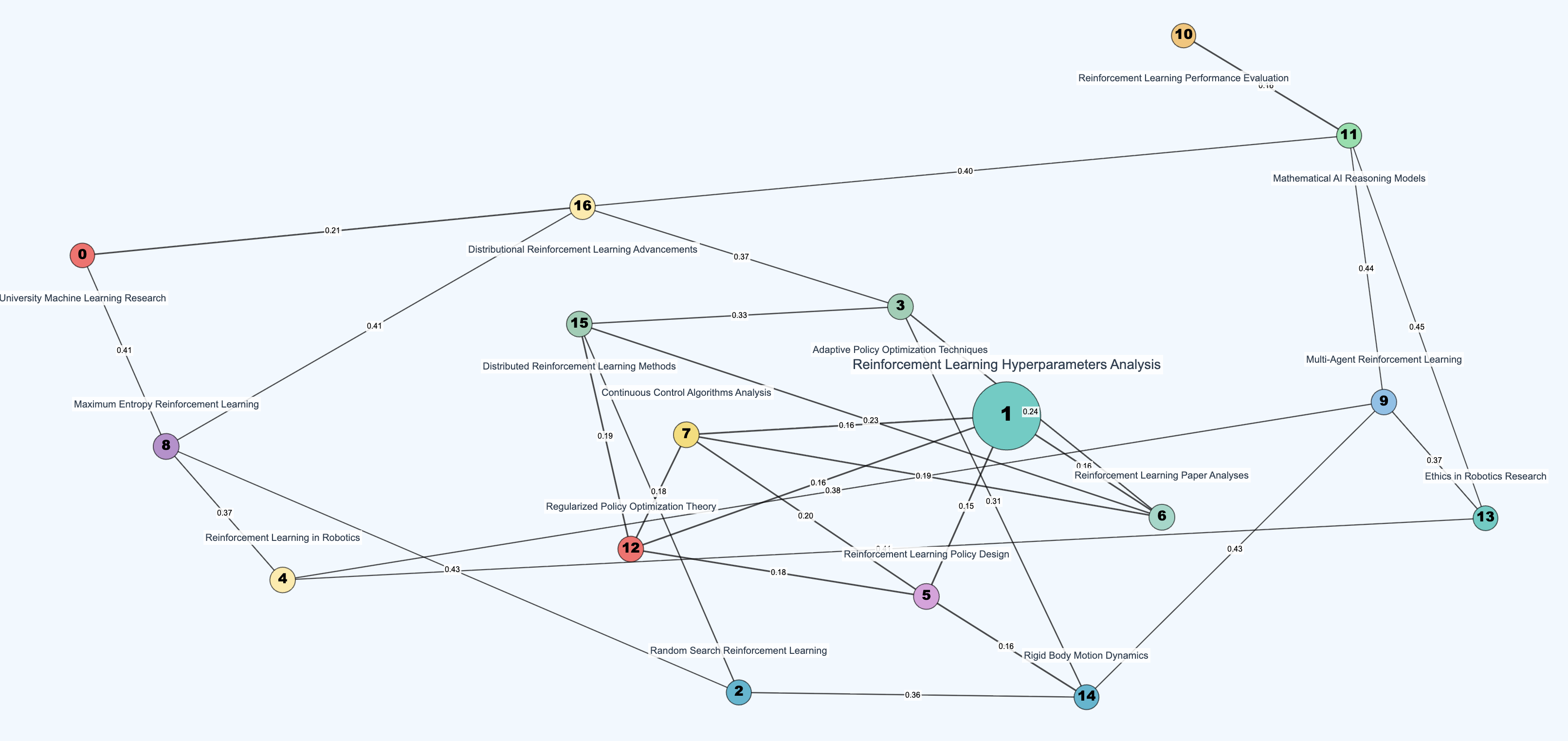}
\caption{Clusters within the RL collection. Nodes represent semantically grouped document segments, showing recurring research topics and methodological patterns.}
\label{fig:rl_clusters}
\vspace{-3mm}
\end{figure*}

\begin{figure*}[h]
\centering
\includegraphics[width=\linewidth]{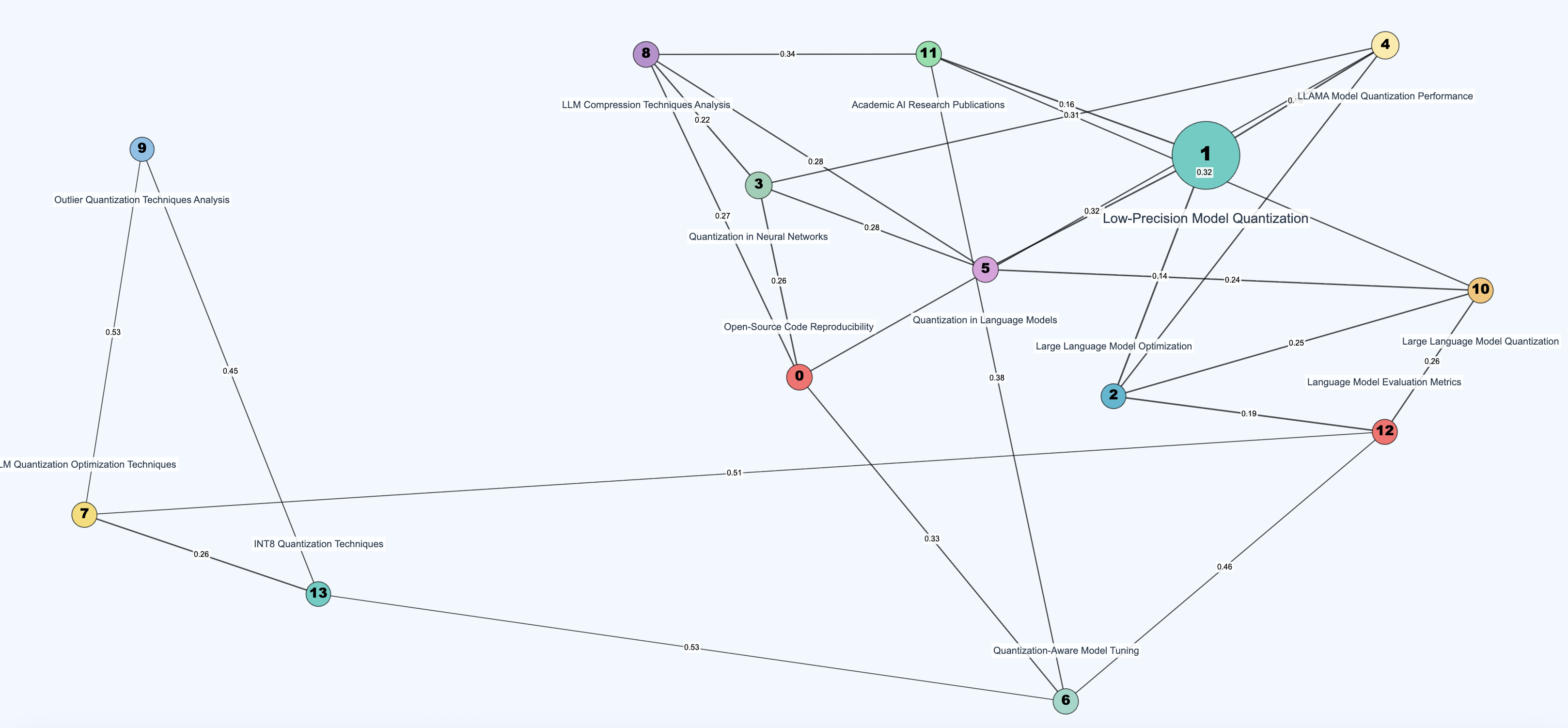}
\caption{Clusters in the Quantization collection. Nodes correspond to document segments, and edges indicate semantic relationships. Clustering highlights both technical and application-specific topics within quantization research.}
\label{fig:llm_healthcare_clusters}
\end{figure*}

\noindent
To better understand the structure and main topics of our document collections, we perform visualization and clustering of document segments. Each collection is first divided into smaller chunks to capture coherent information. Then, we use semantic embeddings to group segments that are similar in meaning. This process allows us to see clusters of related content and the overall thematic organization in each domain. Figures~\ref{fig:hate_speech_clusters}–\ref{fig:llm_healthcare_clusters} illustrate these cluster graphs.

\paragraph{Graph ML Collection}
For the Graph Machine Learning (Graph ML) collection, clustering reveals several key areas of research. Each node in the graph represents a document segment, and edges indicate semantic relationships between them. The main themes identified include:
\begin{itemize}
    \item Performance analysis of different graph network architectures
    \item Techniques for learning node embeddings in graphs
    \item Recent advancements in graph neural networks
    \item Considerations for ethical oversight in graph-based models
    \item Recommendations and applications at web scale
\end{itemize}

\paragraph{Reinforcement Learning Collection}
In the Reinforcement Learning (RL) collection, clustering allows us to identify recurring research directions and methods. The key themes observed are:
\begin{itemize}
    \item Advances in distributional reinforcement learning
    \item Maximum entropy reinforcement learning approaches
    \item Regularized policy optimization techniques
    \item Random search methods in reinforcement learning
    \item Applications of RL in robotics
    \item Analysis of continuous control algorithms
    \item Dynamics of rigid body motion in RL environments
\end{itemize}

\paragraph{Quantization Collection}
The Quantization collection focuses on research related to reducing model size and improving computational efficiency. Clustering shows several distinct themes:
\begin{itemize}
    \item General quantization techniques for neural networks
    \item Analysis of outlier quantization methods
    \item Compression strategies for large language models (LLMs)
    \item Low-precision quantization and its effects
    \item Performance of LLaMA models under quantization
\end{itemize}

\noindent
Overall, these visualizations provide a clear view of how different research themes are distributed across the collections. By examining the clusters, we can identify major areas of focus, connections between topics, and the diversity of content within each domain.

\subsection{Example insights from our method for the interpretability question}

\begin{itemize}
    \item \textbf{Insight 1: Modular Syntax vs Task Layers}

    \textbf{Hook:} Separate layers that learn syntax from layers that learn task details.
    
    \textbf{Body: } Alternate layer roles across depth. Use lower layers to capture broad syntactic structure with low-rank projections, while higher layers handle task-specific reasoning using a small set of high impact attention heads. This makes the model easier to interpret and removes redundant computation.

    \item \textbf{Insight 2: Dynamic Rank and Head Adaptation}

    \textbf{Hook:} Let the model adapt its rank and head usage based on the task.

    \textbf{Body: } Build modules that adjust attention head usage and representation rank to match task complexity. For syntax-light tasks, prune most heads and use compact subspaces. For linguistically rich tasks, keep the essential heads and allow larger rank. Active components will then directly reflect task needs and are easier to inspect.

    \item \textbf{Insight 3: Interpretable Models for Real Time Use}

    \textbf{Hook:} Apply simple, interpretable pruning for on device NLP systems.

    \textbf{Body: } In real time or low resource settings, retain only a few interpretable attention heads that encode syntax, alignment, or position. Pair these with low-rank encodings to make models faster and simpler to analyze. Performance can then be traced to a small number of clear components.

\end{itemize}

\end{document}